\documentclass[sn-mathphys]{sn-jnl}

\usepackage[utf8]{inputenc}
\usepackage[T1]{fontenc}   
\usepackage{hyperref}      
\usepackage{url}           
\usepackage{booktabs}      
\usepackage{amsfonts}      
\usepackage{nicefrac}      
\usepackage{microtype}     
\usepackage{xcolor}        

\usepackage{subcaption}
\usepackage{graphicx}
\usepackage{amsmath}

\usepackage{color, soul}

\jyear{2023}
\theoremstyle{thmstyleone}

\theoremstyle{thmstyletwo}

\theoremstyle{thmstylethree}

\usepackage{xspace}
\newcommand{\HVAE}{HVAE\xspace}
\newcommand{\HVAESR}{EDHiE\xspace}

\begin{document}

\title[Efficient Generator of Mathematical Expressions for Symbolic Regression]{Efficient Generator of Mathematical Expressions for Symbolic Regression}

\author*[1,2]{\fnm{Sebastian} \sur{Me\v{z}nar}}\email{sebastian.meznar@ijs.si}

\author[1]{\fnm{Sa\v{s}o} \sur{D\v{z}eroski}}\email{saso.dzeroski@ijs.si}

\author[3,1]{\fnm{Ljup\v{c}o} \sur{Todorovski}}\email{ljupco.todorovski@fmf.uni-lj.si}

\affil*[1]{\orgdiv{Department of Knowledge Technologies}, \orgname{Jo\v{z}ef Stefan Institute}, \orgaddress{\street{Jamova cesta 39}, \city{Ljubljana}, \postcode{1000}, \country{Slovenia}}}

\affil[2]{\orgname{Jo\v{z}ef Stefan International Postgraduate School}, \orgaddress{\street{Jamova cesta 39}, \city{Ljubljana}, \postcode{1000}, \country{Slovenia}}}

\affil[3]{\orgdiv{Faculty of Mathematics and Physics}, \orgname{University of Ljubljana}, \orgaddress{\street{Jadranska 21}, \city{Ljubljana}, \postcode{1000}, \country{Slovenia}}}

\abstract{
We propose an approach to symbolic regression based on a novel variational autoencoder for generating hierarchical structures, \HVAE. It combines simple atomic units with shared weights to recursively encode and decode the individual nodes in the hierarchy. Encoding is performed bottom-up and decoding top-down. We empirically show that \HVAE can be trained efficiently with small corpora of mathematical expressions and can accurately encode expressions into a smooth low-dimensional latent space. The latter can be efficiently explored with various optimization methods to address the task of symbolic regression. Indeed, random search through the latent space of \HVAE performs better than random search through expressions generated by manually crafted probabilistic grammars for mathematical expressions. Finally, \HVAESR system for symbolic regression, which applies an evolutionary algorithm to the latent space of \HVAE, reconstructs equations from a standard symbolic regression benchmark better than a state-of-the-art system based on a similar combination of deep learning and evolutionary algorithms.}

\keywords{Symbolic regression, Equation discovery, Generative models, Variational autoencoders, Evolutionary algorithms}


\maketitle

\section{Introduction}
Symbolic regression (also known as equation discovery) aims at discovering closed-form equations in collections of measured data~\cite{Schmidt2009DistillingNaturalLaws,Todorovski2017}. Methods for symbolic regression explore the vast space of candidate equations to find those that fit the given data well. They often employ modeling knowledge from the domain of use to constrain the search space of candidate equations. The knowledge is usually formalized into grammars~\cite{Brence2021ProGED} or libraries of model components, such as entities and processes~\cite{Bridewell2008InductiveProcessModeling}. Knowledge-based equation discovery methods have successfully solved practical modeling problems in various domains~\cite{Radinja2021,Simidjievski2020}.

Grammars and libraries of model components are used to generate candidate expressions that might appear in the discovered equations. However, they must be manually crafted, which is a severe obstacle to their broader use. The central aim of this article is to develop a novel generative model of mathematical expressions that can be used for efficient symbolic regression. The model can be trained from a corpus of mathematical expressions from the domain of interest, thus automatically tailoring the space of candidate equations to the application at hand. The developed generative model must have two essential properties to be applicable in such a scenario. First, it should be trainable from a small number of mathematical expressions, e.g., collected from a textbook or from scientific literature in the application domain. Second, the model should encode the expressions in a low-dimensional latent space. The latter space can then be efficiently explored by optimization methods to solve the task of symbolic regression. Lowering the dimensionality of the latent space will significantly increase the efficiency of symbolic regression.

Recently, several variational autoencoders (VAEs) have been shown to be efficient generative models. CVAE~\cite{Gomez2018CVAE} employs a VAE based on recurrent neural networks to encode discrete expressions into a continuous latent space and then decode points from the latent space back into discrete mathematical expressions. This decoder can be used to generate expressions. However, CVAE still generates invalid sequences and requires extensive training data to reduce the likelihood of generating invalid expressions~\cite{Kusner2016Gumbel}. The grammar variational autoencoder, GVAE~\cite{Kusner2017GVAE}, and its successor, SD-VAE~\cite{Dai2018SDVAE}, employ a context-free grammar to ensure the syntactic validity of the generated expressions. Instead of directly training models on sequences, they model the distribution of parse trees that are produced by the grammar while deriving syntactically (and, in the case of SD-VAE, semantically) valid expressions.

We claim that grammars are an unnecessarily powerful and too general formalism for generating mathematical expressions. Grammars add syntactic categories to the expression symbols rendering the parse trees, i.e., the structures modeled with the autoencoder, more complex than the original sequences. This overhead on training expressions inevitably translates to a requirement for more extensive training data and a latent space with larger dimensionality, reducing the efficiency of optimization methods for symbolic regression operating in that latent space.

We propose a novel variational autoencoder for hierarchical data objects, \HVAE, to address these issues. It builds upon the ideas of variational autoencoders for hierarchical data~\cite{Jin2021JTVAE} and gated recursive convolutional neural networks~\cite{Cho2014GatedRCNN}. \HVAE combines simple atomic units with shared weights to encode and decode the individual nodes in the hierarchy. The atomic units are extensions of the standard gated recurrent unit (GRU) cells. The encoding units are stacked into a tree that follows the hierarchy of the training object, and they encode the hierarchy bottom-up, compiling the codes of the descendants to encode the ancestor nodes. The decoding units proceed top-down and use the decoded symbols of the ancestor nodes to decide upon the need to extend the hierarchy with descendant nodes. We claim that \HVAE can be efficiently trained to generate valid mathematical expressions from a training set of modest size, while using a low-dimensional latent space.

\begin{figure}[h!]
  \centering
  \includegraphics[width=.8\linewidth]{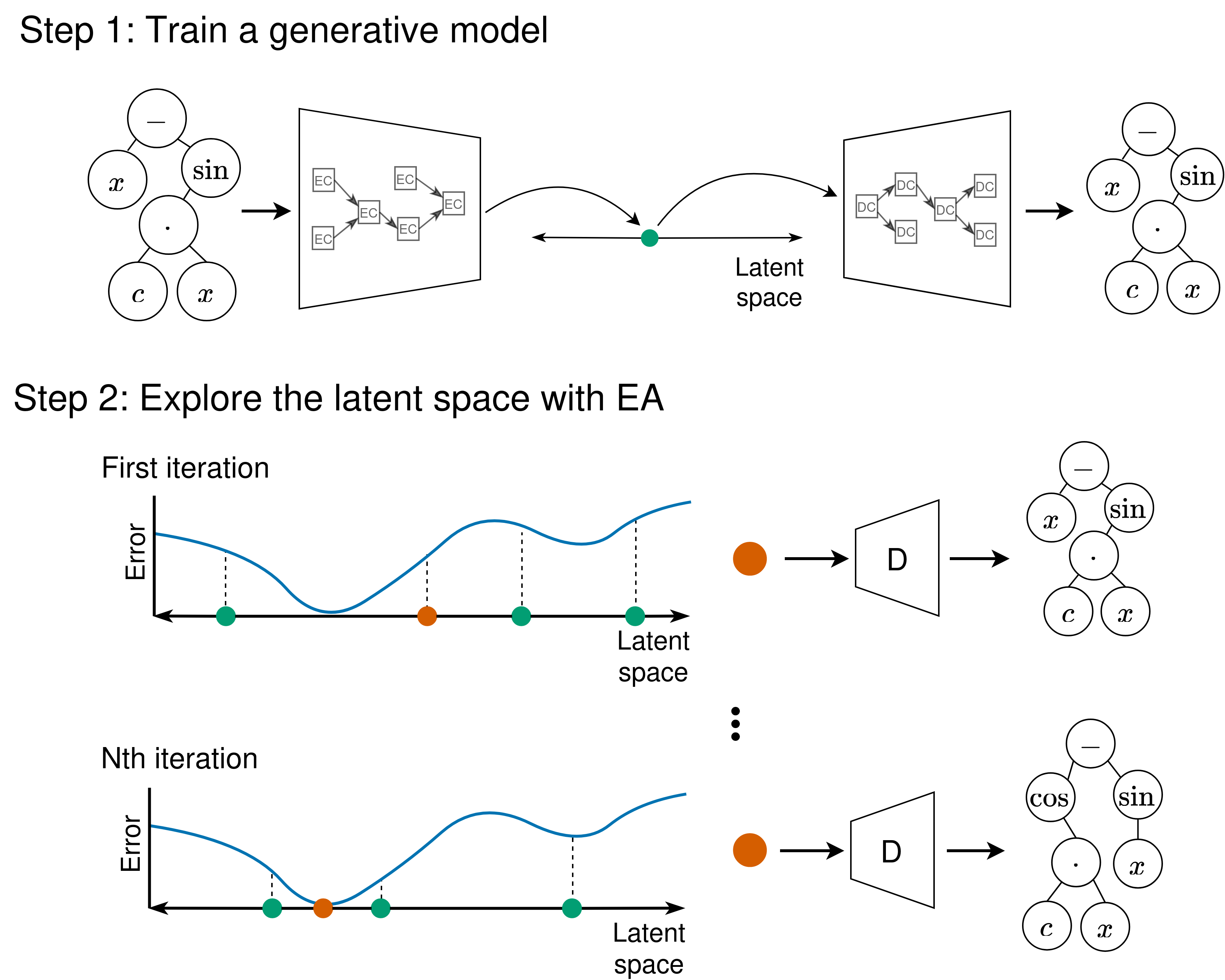}
  \caption{A schematic representation of the \HVAESR approach. In the first step, we train a \HVAE model. In the second step, we explore the latent space of the \HVAE model with an evolutionary algorithm. The red dot represents the best expression in a given iteration.}
  \label{fig:overview}
\end{figure}

We exploit these expected properties of our \HVAE to implement a novel approach for symbolic regression, \HVAESR. It performs an evolutionary search through the latent space of a \HVAE trained on mathematical expression trees as shown in Figure~\ref{fig:overview}. The genetic operations utilize the \HVAE encoder to obtain the expressions' latent codes, generate new individuals with crossover and mutation in the latent space, and decode the latter back to mathematical expressions. \HVAESR can then evaluate the fit of the obtained expressions against the measurements. We conjecture that the performance of \HVAESR on standard benchmarks~\cite{Uy2011,Tegmark2020Feynman} would compare favorably to that of a state-of-the-art symbolic regression methods~\cite{Petersen2021NeuralGP}. The results of our empirical evaluation of \HVAE and \HVAESR confirm our conjectures. \HVAE can achieve better reconstruction of the training expressions with order-of-magnitude fewer training examples while using latent spaces with fewer dimensions. \HVAESR outperforms alternative methods for symbolic regression on the task of reconstructing the ten equations in the Ngyuen benchmark.

We can summarize the contributions of this work as follows:
\begin{itemize}
    \item We propose HVAE, a variational autoencoder for hierarchical data, that can be efficiently trained to generate mathematical expressions from modest amounts of data, while using a low-dimensional latent space.

    \item We introduce \HVAESR, a symbolic regression approach that exploits \HVAE to efficiently search through the space of candidate equations.
\end{itemize}

The remainder of the paper is organized as follows. Section~2 reviews related work on generative models and symbolic regression. We introduce the hierarchical variational autoencoder \HVAE and the symbolic regression approach \HVAESR in Section~3. Section~4 presents the results of the empirical evaluation of \HVAE and \HVAESR. Finally, Section~5 summarizes and discusses the contributions of the presented work and outlines directions for further research.

\section{Related work}
Most of the early successful applications of generative models have been in the domains of text, speech, images, and video, i.e., they have been mainly used for generating unstructured data objects composed of continuous data elements. The discrete data structures that generative models have most often tackled are strings or sequences of characters, where the data elements are discrete symbols. The models that use strings as input (and output) usually do so by training a recurrent neural network~\cite{Sherstinsky2020RNNs}, most commonly using seq2seq autoencoders~\cite{Sutskever2014Seq2seq}.

A major problem of sequence-to-sequence autoencoders is that they do not guarantee the syntactic correctness of the generated expressions. One way to solve this problem is to learn an additional validation model for checking the correctness of the generated sequence~\cite{Janz2017Validator}. Grammar variational autoencoders (GVAE)~\cite{Kusner2017GVAE} use context-free grammars for specifying the space of valid structured data objects. Each data object can be then represented as a sequence of grammar productions (rewrite rules) that derives it. In turn, GVAE encode sequences of rewrite rules that derive objects instead of the objects themselves. The structure of the decoder is constrained to generate valid sequences of rewrite rules that are then used together with the grammar to generate valid expressions.

Dai et al.~\cite{Dai2018SDVAE} propose the use of attribute grammars, i.e., context-free grammars that attach attributes to the grammar's syntactic categories. By prescribing properties and relationships between the attributes, such grammars can also encode semantic constraints on the derived data objects. The attribute grammars, together with SD-VAE, i.e., syntax-directed VAE, can generate expressions that are consistent with a set of both syntactic and semantic constraints. Alternative generalizations of grammars have been used for generative modeling of program source code in high-level languages~\cite{Bielik2016PHOG}.

Most of the above approaches can also generate mathematical expressions. However, they need the complex formalism of grammars to generate more complex data structures, most often molecular structures~\cite{Gomez2018CVAE, Dai2018SDVAE}. Since mathematical expressions can be represented as simpler structures, i.e., binary trees, our work concerns generative models for hierarchical (tree-structured) data.

Hierarchical data have been tackled by generative models in several ways. By making a node depend on its parent and previous sibling, DRNN~\cite{Alvarez2017TreestructuredDW} combines representations obtained from the depth-wise and width-wise recurrent cells to generate new nodes, which proves useful for recovering the structure of a tree. On the other hand, Tree-LSTM~\cite{tai2015TLSTM} and JT-VAE~\cite{Jin2021JTVAE} focus on adapting equations for recurrent cells to encode (and decode) hierarchical structures more efficiently. Tree-LSTM proposes a generalization of the LSTM cell for encoding trees into a representation that proves effective for classification tasks and semantic relatedness of sentence pairs. JT-VAE adapts recurrent cells for tree message passing. Trees are used as scaffolding for the graph that represents molecules. Encoding and decoding are thus split into four parts: encoding of the graph, encoding of the tree, decoding of the tree, and decoding of the graph. While these adaptations are similar to the ones presented in our work, their focus is on encoding more general structures that are unrelated to mathematical expressions.

Note that our model falls into the general framework of gated \emph{recursive} convolutional neural networks~\cite{Cho2014GatedRCNN} that combine atomic units with shared parameters in a hierarchy. The output of the root node produces a fixed-length encoding of a data object with an arbitrarily complex structure. Another model, marginally related to ours, is the one of equivalence neural networks~\cite{Allamanis2017SemanticExpressions}. The encoding produced by these networks follows the expressions' semantic similarity and equivalence, in contrast to their syntactic similarity, which is followed by all the other approaches, including ours.

Finally, our work is also related to algorithms for equation discovery and symbolic regression. Most of them generate candidate expressions for equations first and then estimate the values of their constant parameters by matching the equations against data in the second phase. Classical symbolic regression approaches~\cite{Schmidt2009DistillingNaturalLaws, Guimera2020BayesianMachineScientist, cranmer2023pysr}, based on evolutionary algorithms, use stochastic generators of expression trees: At the beginning, the expression trees are randomly sampled, and later on, they are transformed using the evolutionary stochastic operations of mutation and cross-over. In contrast, process-based modeling approaches~\cite{Bridewell2008InductiveProcessModeling} generate equations by following domain-specific knowledge (provided by the user) that specifies a set of entities (variables) and processes (interactions among entities). Grammar-based approaches to equation discovery employ user-specified context-free grammars (which can also be based on domain knowledge), deterministic~\cite{Todorovski1997BiasED} or probabilistic~\cite{Brence2021ProGED}, as efficient generators of expressions.
 
Recently, many symbolic regression approaches based on neural networks have been proposed~\cite{Tegmark2020Feynman, Martius2016EQL, Biggioetal21, dAscoli2022deep, Kamienny2022transformers, Petersen2021CharGen}. In particular, Deep Symbolic Optimization, DSO approaches symbolic regression (among other optimization tasks~\cite{Petersen2021NeuralGP}) by combining neural networks and reinforcement learning with evolutionary algorithms. The neural networks are used to sample the individuals in the initial population of the evolutionary algorithm and are retrained at each iteration to focus on expressions leading to better fit. It is closely related to our work, since it combines similar methods. Yet our focus here is on efficient neural networks for generating mathematical expressions that are trained before the beginning of the evolutionary process.

\section{Methodology}
We start this section by briefly introducing the task of symbolic regression and the search space of mathematical expressions (Section ~\ref{sec:sym-reg}). After this, we introduce variational autoencoders and the structure of the hierarchical variational autoencoder, \HVAE (Section~\ref{sec:VAEs}). We finish the section by explaining how to use \HVAE for generating mathematical expressions and how to combine it with an evolutionary algorithm for symbolic regression (Section~\ref{sec:generation-hvae}).

\subsection{Symbolic regression and expression trees}\label{sec:sym-reg}
Symbolic regression (SR) is the machine learning task of discovering equations in collections of measured data. Symbolic regression methods take a data set $S$ consisting of multiple measurements of a set of real-valued variables $V = \{ x_1, x_2, \ldots, x_p, y \}$, where $y$ is a designated target variable. The output of SR is an equation of the form $y = f(x_1, x_2, \ldots, x_p)$, where the right-hand side of the equation is a closed-form mathematical expression. The equation should provide an optimal fit against the measurements from $S$, i.e., minimize the discrepancy between the observed values of the target variable $y$ and values calculated by using the equation. Symbolic regression methods usually follow the parsimony principle, preferring simpler expressions over more complex ones.

Symbolic regression methods search through the space of candidate mathematical expressions for the right-hand side of the equation to find the one that optimally fits the measurements. Mathematical expressions can be represented in different ways. We commonly use the infix notation, where operators are placed between two sub-expressions they operate on, e.g., $A + B$, where $A$ and $B$ are sub-expressions. Infix notation uses parentheses to indicate the order in which the operations need to be performed. Prefix (Polish) or postfix (reverse Polish) notations do not need parentheses since the operators are written before or after the two sub-expressions, e.g., $+ A B$ or $A B +$. The three notations correspond to different traversals of the nodes in an expression tree. The latter is a hierarchical data structure, where the inner nodes correspond to mathematical operators and functions, while the leaf nodes correspond to variables and constants.

\begin{figure}[h!]
  \centering
  \includegraphics[width=0.7\linewidth]{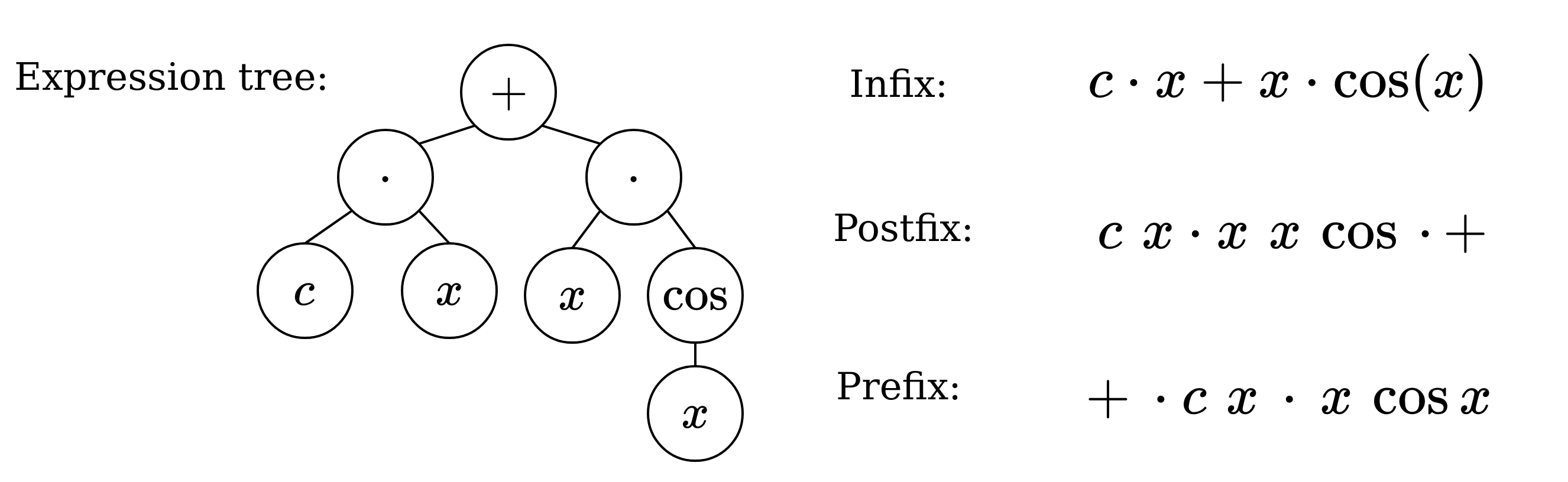}
  \caption{An expression tree with a height of four and three sequence-based representations of the corresponding mathematical expression.}
  \label{fig:eq-representation}
\end{figure}

In symbolic regression, the constants' values are fitted against the measured data from $S$, while variables include elements from $V$ without the target variable. We assume binary expression trees since standard arithmetic operators are binary. We take that the second descendant node is null in the inner nodes corresponding to single-argument functions. We define the height of an expression tree as the number of nodes on the longest path from the root node to one of the leaves. Figure~\ref{fig:eq-representation} depicts an example expression tree with a height of four, along with the corresponding mathematical expression in different notations.

Our model generates expression trees, as they have several advantages over sequences (strings). Firstly, it is easy to achieve syntactic correctness, since operators and functions are in the inner nodes, while variables and constants are in the leafs. Secondly, information needs to travel at most $\log n$ steps up the tree (up to the tree's height) instead of $n$ steps along the sequence (up to the length of the sequence). Lastly, sub-expressions can be encoded independently of each other during the encoding process.

\subsection{Hierarchical variational autoencoder}\label{sec:VAEs}
In recent years, variational autoencoders~\cite{Kingma2014VAE} have emerged as one of the most popular generative models. The reason for this is that, when trained correctly, variational autoencoders map the observed data with an unknown distribution into a latent representation with a known distribution. This results in a continuous latent space, from which one can sample and synthesize new data. In contrast to a (deterministic) autoencoder, where the encoder outputs a latent representation $z$ that is directly fed into the decoder, the encoder in the variational autoencoder outputs the parameters for an approximate posterior distribution, e.g., $\mu$ and $\sigma$ in the case of a latent space parameterized by a multivariate Gaussian distribution.

Thus, a representation $z$ that is fed into the decoder is sampled from the underlying distribution with the learned parameters ($\mu$, $\sigma$). The loss of the variational autoencoder is the reconstruction error, i.e., the difference between the input to the encoder and the output of the decoder. Additionally, variational autoencoders typically use Kullback-Leibler (KL) divergence~\cite{Kullback1951KL} as the regularization term for the loss. The loss can thus be calculated as:
\begin{equation}
    J(x, \mu_z, \sigma_z) = J_{\text{rec}}(x) + \lambda\cdot \textsc{KL}(\mathcal{N}(\mu_z,\sigma_z) \, \| \, \mathcal{N}(0, I)),
\end{equation}
where $J_{\text{rec}}(x)$ is the reconstruction loss of $x$ and $\lambda\geq 0$ the regularization cost parameter. In case the underlying distribution is Gaussian, KL divergence to an isotropic unit Gaussian can be estimated as 
\begin{equation}
    \textsc{KL}(\mathcal{N}(\mu_z,\sigma_z) \, \| \, \mathcal{N}(0, I))=\frac{1}{2}(1 + \log \sigma^2_z - \mu^2_z - \sigma^2_z). 
\end{equation} 
We use cost annealing~\cite{Bahuleyan2018natural} to focus on the reconstruction error (i.e., use small values of $\lambda$) at the beginning and then gradually shift the focus towards the smoothness of the latent space by increasing the value of $\lambda$.

\subsubsection{Model overview}
Our approach uses a variational autoencoder structure that consists of an encoder and a decoder. The encoder takes tree-structured data as input and outputs a distribution in the latent vector space, represented with the mean ($\mu_z$) and the logarithm of the variance ($\log\sigma_z$) vectors. The decoder works in the opposite direction, sampling a point from the latent vector space as input and transforming it into a binary expression tree. To make the backward propagation possible, we sample points with the reparametrization trick.

\begin{figure*}
     \centering
     \resizebox{\linewidth}{!}{
     \begin{subfigure}[b]{0.49\textwidth}
         \centering
         \includegraphics[width=\textwidth]{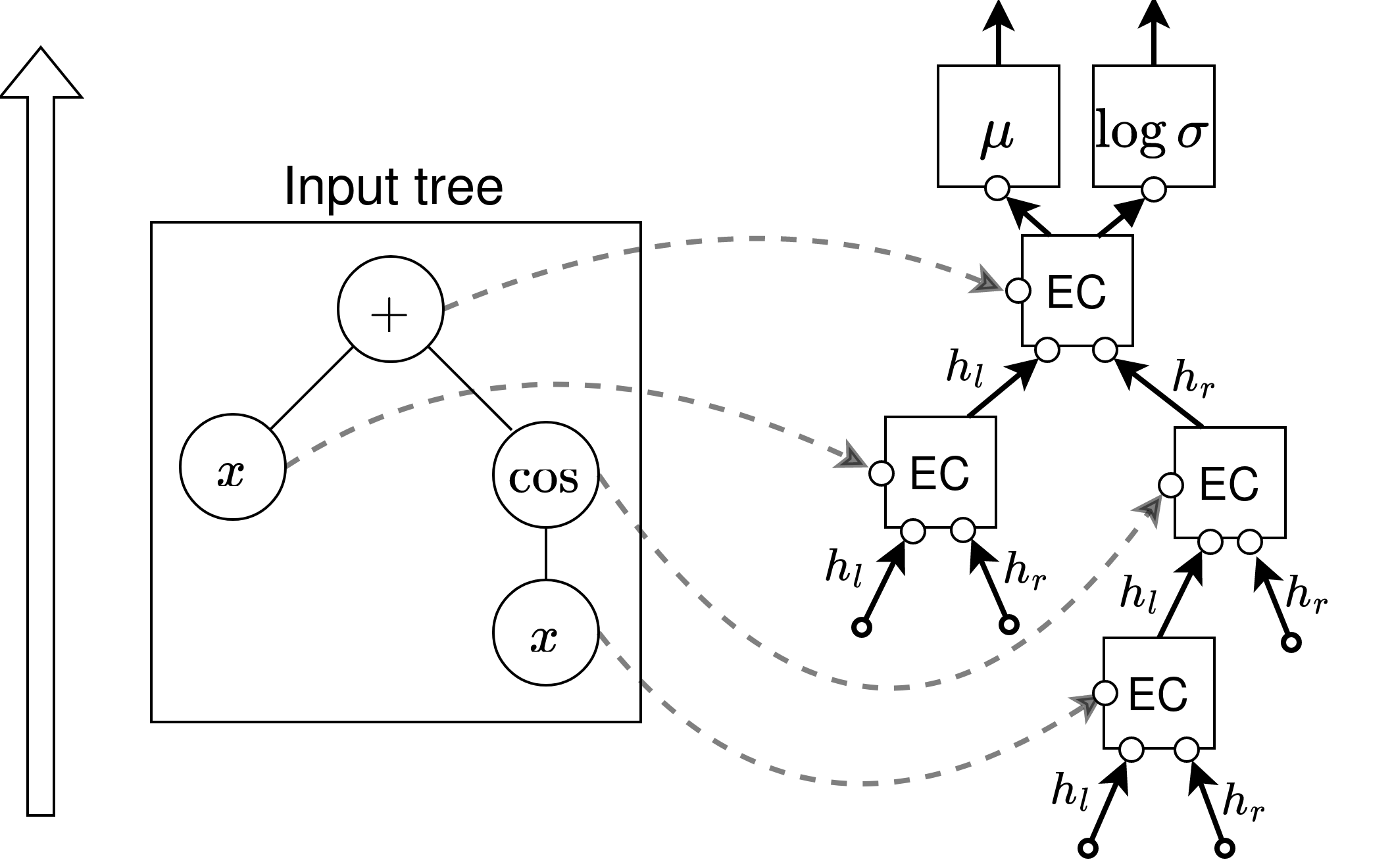}
         \caption{Encoding}
         \label{fig:encoding}
     \end{subfigure}
     \hfill
     \begin{subfigure}[b]{0.49\textwidth}
         \centering
         \includegraphics[width=\textwidth]{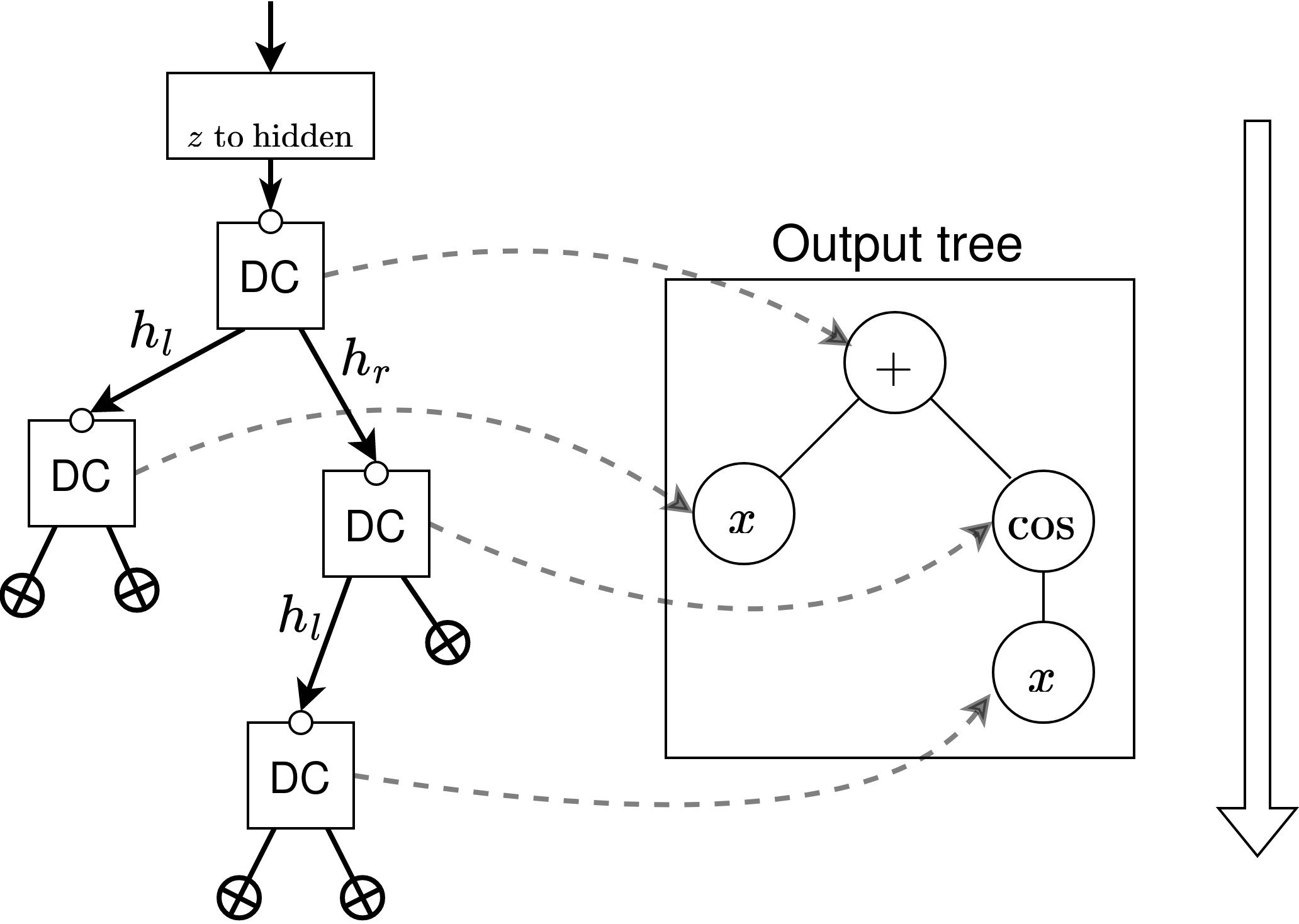}
         \caption{Decoding}
         \label{fig:decoding}
     \end{subfigure}
     }
    \caption{The processes of (a) encoding and (b) decoding the expression tree of $x+\cos x$. The acronyms EC and DC stand for ``encoding cell'' (introduced in Section~\ref{sec:encoder}) and ``decoding cell'' (introduced in Section~\ref{sec:decoder}).}
\end{figure*}

Trees are encoded recursively, starting from leafs and ending at root nodes. To encode a subtree with a root in $n$, we first encode its left and right subtrees. We then pass their codes, along with the symbol in the node $n$, as inputs to the encoding cell (further described in Section~\ref{sec:encoder}). This cell outputs the code of the subtree rooted in $n$. At the beginning of the recursion, in each leaf node, the codes corresponding to the (missing) children are assumed to be vectors of zeros. Once the root of the tree is encoded, its code is passed through two fully connected layers that give the mean and log-variance vectors that form the latent representation of the tree. Figure~\ref{fig:encoding} illustrates the recursive encoding process on the expression $x+\cos x$.

The first layer of the decoder transforms the sampled point from the latent space into the code of the hierarchy. After this, the tree is generated recursively by passing the code of the current node (subtree) through the decoding cell (further described in section~\ref{sec:decoder}). This cell takes the code of the node (subtree) as input and generates a node symbol, along with the codes of the two child nodes. There are three possible symbol types. If we encounter an operator, both child nodes are generated recursively. On the other hand, if the symbol represents a function, we only generate the left child. Lastly, if the symbol is either a variable or a constant, no further child nodes are generated in this branch. This process is shown in Figure~\ref{fig:decoding}, where the expression $x+\cos x$ is decoded.

During training, we follow the structure of the encoded tree and try to predict the correct node symbols. In turn, we jointly learn to predict the structure of the expression tree and the symbols inside the node, since the structure is determined by the symbols. We calculate the loss using cross-entropy on a sequence of symbols obtained with the in-order traversal of the expression tree. Some additional implementation details are explained in Appendix~\ref{app:implementation}.

\subsubsection{Encoder}\label{sec:encoder}
The encoding proceeds in two phases. The first follows the hierarchy of the input and applies the encoding cell to each node of the hierarchy as described above. In the second phase, the code of the root node is transformed into the mean and log-variance vectors of the input's latent representation.

Encoding comprises a GRU21 cell, which we have adapted from the GRU cell~\cite{Cho2014GRU}. The (output) code $h$ in GRU21 is computed from the input vector $x$, and codes $h_l$ of the left and $h_r$ of the right child with the following equations:
\begin{gather}
    r = \varphi_S (W_{ir}x + b_{ir} + W_{hr}(h_l \concat h_r) + b_{hr})\label{eq:r-enc}\\
    u = \varphi_S (W_{iu}x + b_{iu} + W_{hu}(h_l \concat h_r) + b_{hu})\label{eq:z-enc}\\
    n = \textrm{tanh} (W_{in}x + b_{in} + r*(W_{hn}(h_l \concat h_r) + b_{hn}))\label{eq:n-enc}\\
    h = (1-u)*n + \frac{u}{2}*h_l + \frac{u}{2}*h_r,\label{eq:h-enc}
\end{gather}
where $\varphi_S$ denotes the Sigmoid activation function. In these expressions, $r$, $u$, and $n$ represent the standard reset gate, update gate and candidate activation vectors from a GRU cell. When compared to the original equations of the GRU cell, Equations~\eqref{eq:r-enc},\eqref{eq:z-enc},\eqref{eq:n-enc} exhibit two differences. First, instead of the code of the previous symbol in the sequence, the concatenation of the codes $h_l$ and $h_r$ of the child nodes is used (denoted by $(h_l \concat h_r)$). Second, the dimension of the weight matrices $W_{hr}, W_{hu}, W_{hn}$ must be $\dim(h_l)+\dim(h_r)$ instead of $\dim(h)$. Thus, while Equation~\eqref{eq:h-enc} remains similar to the original one, we change the second term (from its usual form $u*h_{t-1}$) to $\frac{u}{2}*h_l + \frac{u}{2}*h_r$, to retain information from the codes of the two child nodes. Recall that $*$ denotes the element-wise multiplication of vectors.

In the second phase, the model transforms the code of the root node into the latent representation of the input expression through two fully-connected layers.

\subsubsection{Decoder}\label{sec:decoder}
The decoding also comprises two phases. In the first, a fully-connected layer transforms a point from the latent vector space into the code of the root node. In the second phase, the decoding cell is recursively deployed to decode each of the nodes in the expression tree.

\begin{figure}[h!]
 \centering
 \includegraphics[width=0.3\linewidth]{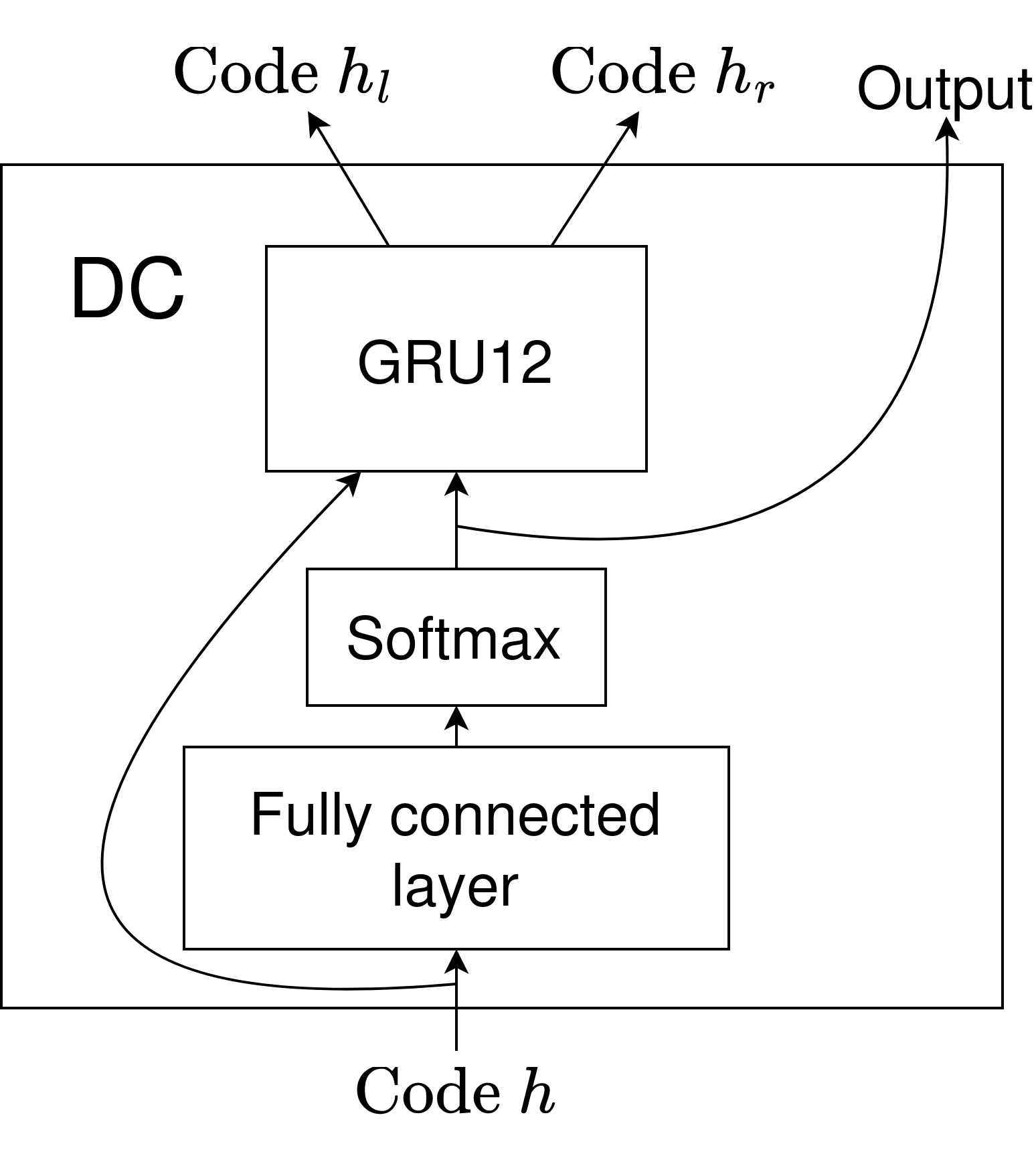}
 \caption{The structure of the decoding cell.}
 \label{fig:de-cell}
\end{figure}

Figure~\ref{fig:de-cell} depicts the structure of the decoding cell. The cell is composed of a fully connected layer, a softmax layer, and the GRU12 cell, an adaptation of the original GRU cell. The input code is first passed through the sequence of a fully-connected and a softmax layer. The latter creates the vector of probabilities, from which the most probable output symbol is chosen. If the output symbol is either a constant or a variable, the decoding stops. Otherwise, the output vector is also used as an input to the GRU12 cell, together with the code that is given as input into the decoding cell. The GRU12 cell produces two codes, one for the left and one for the right child.

GRU12 computes the two codes $h_l$ and $h_r$ for the child nodes using the input vector $x$ and the code $h$ with the following equations:
\begin{gather}
    r = \varphi_S (W_{ir} x + b_{ir} + W_{hr} h + b_{hr}) \label{eq:r-dec} \\
    u = \varphi_S (W_{iu} x + b_{iu} + W_{hu} h + b_{hu}) \label{eq:z-dec}\\
    n = \textrm{tanh} (W_{in} x + b_{in} + r * (W_{hn} h + b_{hn})) \label{eq:n-dec} \\
    d = (1-u) * n + u * (h \concat h) \label{eq:l-dec} \\
    d \equiv h_l \concat h_r \label{eq:l-split}
\end{gather}
There are two major differences between GRU12 and the original GRU cell. First, the vectors $r$,$u$, and $n$ in Equations~\eqref{eq:r-dec},\eqref{eq:z-dec},\eqref{eq:n-dec} are of dimension $2\cdot\dim(h)$ instead of $\dim(h)$. Consequently, all bias vectors are of dimension $2\cdot\dim(h)$, and all weight matrices have an output dimension of $2\cdot\dim(h)$. Second, in Equation~\eqref{eq:l-dec}, the code $h$ is concatenated with itself to make the dimensions in the equation match. Vector $d$ is then split in half in Equation~\ref{eq:l-split}. The first part is used as a code for the left child, while the second is used as a code for the right child.

\subsection{Generating expressions for symbolic regression}\label{sec:generation-hvae}
Recall that the goal of symbolic regression is to efficiently search through the space of mathematical expressions and find the one that, when used on the right-hand of an equation, fits given measurements well. In this section, we explain how to use \HVAE for generating expressions.

\subsubsection{\HVAE as a generative model}\label{sec:sampling}
We can generate expressions in two ways, corresponding to two different symbolic regression scenarios. The first way, which aims at discovering equations from data, samples random vectors from the standardized Gaussian distribution $\mathcal{N}(0, I)$ in the latent space and passes them through the decoder.

On the other hand, we might want to generate expressions in a scenario that corresponds to the revision of existing equations to fit newly gathered data. Here, we want to generate mathematical expressions that are similar to the one given as input. Our approach achieves this by encoding an expression and sampling its immediate neighborhood in the latent space. We expect these points to be decoded into expressions similar to the one given as input. We will show that \HVAE meets this expectation in Section~\ref{sec:eval-hvae-li}.

\subsubsection{Evolutionary algorithm operators}\label{sec:EA}
Finally, we can search the latent space spanned by our model with evolutionary algorithms~\cite{Koza1994}, one of the most commonly used paradigms for symbolic regression. Evolutionary algorithms explore the search space by first randomly sampling individuals for the initial population. Then they repetitively generate new populations by combining pairs of individuals from the current population with the genetic operators of mutation and crossover.

An \emph{individual} in a population is in our case a real-valued vector $z$, corresponding to the code of an expression tree in the latent vector space. Using the \HVAE model, $z$ can be decoded into an expression tree. To calculate the individual's fit against the training data, we first fit the values of the constant parameters in the decoded expression tree and then measure the error of the equation with the resulting expression on the right-hand side (with respect to the training data). We generate the \emph{initial population} by randomly sampling individuals from the Gaussian distribution $\mathcal{N}(0, I)$.

\emph{Crossover} combines two individuals, referred to as parents $z_A$ and $z_B$, into an offspring $z_O$. We generate the latter as a convex combination of $z_A$ and $z_B$, i.e. $z_{\text{O}}= (1-a) \cdot z_A + a \cdot z_B$, where $a$ is sampled from the uniform distribution on the interval $[0, 1]$. For values of $a$ close to $0$ and $1$, the offspring is close to one of the parents, while values of $a$ close to $0.5$ lead to an offspring equally dissimilar to both parents.

The \emph{mutation} operator transforms an individual $z$ into a mutated individual $z_M$. To perform a mutation, we first decode $z$ into an expression tree and immediately encode it back into its latent space representation to obtain the value of $\sigma_z$. Now, we can mutate $z$ into an individual with a syntactically similar expression by sampling from $\mathcal{N}(\mu_z, \sigma_z)$ or into a random individual by sampling the offspring $z_O$ from $\mathcal{N}(0, I)$. Similarly to the case of crossover, we interpolate between these two extremes by sampling from $\mathcal{N}(a \cdot \mu_z + (1 - a) \cdot 0, a \cdot \sigma_z + (1 - a) \cdot I) = \mathcal{N}(a \cdot \mu_z, a \cdot \sigma_z + (1 - a) \cdot I)$, where $a$ is randomly sampled from the uniform distribution on the interval $[0,1]$. When $a$ is close to $0$, the offspring $z_O$ is chosen at random (see the first paragraph of Section~\ref{sec:sampling}). On the other hand, when $a$ is close to $1$, $z_O$ is syntactically similar to $z$ (second paragraph of Section~\ref{sec:sampling}).

We implement the \HVAESR (Equation Discovery with Hierarchical variational autoEncoders) approach for symbolic regression by combining \HVAE with evolutionary algorithms using these operators. Our implementation uses pymoo~\cite{pymoo} for evolutionary algorithms and ProGED~\cite{Brence2021ProGED} functionality for evaluating the fit of a candidate equation.

\section{Evaluation}
In this section, we will investigate the validity of our hypothesis that the hierarchical variational autoencoder is a more efficient generator of mathematical expressions than the alternative VAEs for sequences by conducting two series of computational experiments. In the first series, we are going to evaluate the performance and efficiency of \HVAE on the task of generating mathematical expressions. In the second series, we will evaluate the performance of \HVAESR on the symbolic regression downstream task.

\subsection{The performance of \HVAE}
We start this section by introducing the experimental setup (Section~\ref{sec:eval-hvae-setup}). We continue with reporting the experimental results of evaluating \HVAE with respect to the reconstruction error (Section~\ref{sec:eval-hvae-re}), efficiency in terms of the size of training data needed, the dimensionality of the latent space (Section~\ref{sec:eval-hvae-eff}), and finally the smoothness of the latent space (Section~\ref{sec:eval-hvae-li}). In Appendix~\ref{app:li}, we further justify our claim that points close in the latent space of \HVAE are decoded into similar expressions.

\subsubsection{Experimental setup}\label{sec:eval-hvae-setup}
\textbf{Data sets.} We estimate the reconstruction error of the variational autoencoders on a collection of six synthetic data sets, ranging from small ones, including simple expressions, to large ones, including complex expressions. The data sets are as follows:
\begin{description}
    \item[AE4-2k, AE5-15k, and AE7-20k] have 2, 15, and 20 thousand mathematical expressions with trees with a maximum height of four, five, and seven. These expressions can contain constants, variables, and the operators $+$,$-$,$\cdot$,$/$, and $\hat{\mkern6mu}$.
    \smallskip
    \item[Trig4-2k, Trig5-15k, and Trig7-20k] are the same as above, but the expressions also contain the sine and cosine functions.
\end{description}

We create these data sets with the ProGED~\cite{Brence2021ProGED} system by randomly sampling mathematical expressions from a given probabilistic context-free grammar. The generated expressions are simplified using the Python library SymPy~\cite{Meurer2017SymPy}. The context-free grammars that constrain the output of GVAE and the ones used to generate the data sets are documented in Appendix~\ref{app:grammars}.

\smallskip
\noindent
\textbf{Parameter setting}. We train GVAE and CVAE for $150$ epochs with the following values of their hyper-parameters: latent dimension $=128$, hidden dimension $=128$, batch size $=64$, kernel sizes of the convolution layers $= 2,3,4$, and the ADAM optimizer~\cite{Diederik2015Adam}. For reconstruction results created with our approach (\HVAE), the hyper-parameters are: latent size $=128$, batch size $=32$, and the ADAM optimizer with the default learning rate. For the first 1,800 iterations $i$, we calculate the regularization cost parameter $\lambda$ using $\lambda_i = 0.5\cdot(\tanh \frac{i-4,500}{2}+1)$, after this, we set $\lambda_i$ to $\lambda_{1,800}$.

\smallskip
\noindent
\textbf{Estimating the reconstruction error}. The Levenshtein distance~\cite{Levenshtein1965BinaryCC} (often referred to as the edit distance) quantifies the dissimilarity of two strings in terms of the number of insertion, removal, and substitution operations that are needed to transform one string into the other. We use this distance to test how well our autoencoder recreates expressions. 

We first pass the expression through the VAE to get the predicted expression. If needed, we validate the syntactical correctness of the latter and transform it into an expression tree. We then traverse the input and the output trees in post-order (left child, right child, node symbol) to obtain the input and the output expressions in the postfix notation (which does not require parentheses and is hence more suited for calculating the distance between expressions). Finally, we calculate the edit distance between those two strings.

To estimate the reconstruction error on unseen expressions, we use five-fold cross-validation with the same splits across all methods. GVAE and CVAE sometimes produce invalid expressions, which we discard from the evaluation. Because of this, the results in Section~\ref{sec:eval-hvae-re} and~\ref{sec:eval-hvae-eff} might be biased in favor of CVAE due to many syntactically incorrect expressions being discarded. Note that GVAE has fixed-size input (and output) that might be too short for encoding all the grammar rules needed to derive an expression. In those cases, GVAE returns empty strings, which we consider invalid expressions. CVAE, on the other hand, produces syntactically incorrect expressions such as $x c (/x)c)\sin \sin ( c ) )$, $\cdot\cdot x-c\cdot /\sin(x))$, or $(/x(-x)c)$ (presented here in infix notation).

\subsubsection{Out-of-sample reconstruction error}\label{sec:eval-hvae-re}
Table~\ref{tab:valid} compares the out-of-sample reconstruction error and the ratio of invalid expressions for the three variational autoencoders. Our hierarchical VAE significantly outperforms the other two methods on all data sets. An interesting observation is that GVAE works consistently better on expressions involving trigonometric functions, while \HVAE and CVAE perform worse. The reason for the opposite effect is probably the following: for GVAE, functions only represent yet another production rule in the grammar, while for \HVAE and CVAE they drastically change the structure of the expression (tree). This translates to better performance of GVAE, as expressions with trigonometric functions are usually shorter, given that the nodes corresponding to the trigonometric functions have only one descendant instead of the usual two.

\begin{table}[h]
	\caption{The out-of-sample reconstruction error and the percentages of syntactically incorrect expressions generated by the three variational autoencoders.}
	\label{tab:valid}
    \centering
    \resizebox{\linewidth}{!}{
	\begin{tabular}{lcccccc}
    \toprule
    & \multicolumn{2}{l}{HVAE}          & \multicolumn{2}{l}{GVAE}                & \multicolumn{2}{l}{CVAE}        \\ \midrule
    Dataset & Edit distance & Invalid & Edit distance & Invalid & Edit distance & Invalid \\\midrule
    AE4-2k    & \bfseries 0.076 ($\pm$ 0.024) & \bfseries 0.0 ($\pm$ 0.0)& 3.959 ($\pm$ 0.135)& 0.2 ($\pm$ 0.0)  & 3.873 ($\pm$ 0.132) & 33.8 ($\pm$ 1.1)\\
    Trig4-2k  & \bfseries 0.119 ($\pm$ 0.026) &\bfseries 0.0 ($\pm$ 0.0)& 3.199 ($\pm$ 0.068)& \bfseries 0.0 ($\pm$ 0.0)& 3.619 ($\pm$ 0.045) & 48.3 ($\pm$ 0.6)\\
    AE5-15k   & \bfseries 0.079 ($\pm$ 0.014)&\bfseries 0.0 ($\pm$ 0.0) &2.827 ($\pm$ 0.280) & $<$ 0.1 ($\pm$ 0.0) & 1.547 ($\pm$ 0.466) & 3.5 ($\pm$ 0.0) \\
    Trig5-15k & \bfseries 0.093 ($\pm$ 0.010) &\bfseries 0.0 ($\pm$ 0.0) &1.489 ($\pm$ 0.195)& $<$ 0.1 ($\pm$ 0.0)& 2.086 ($\pm$ 0.346)& 13.9 ($\pm$ 0.0)\\
    AE7-20k   & \bfseries 0.501 ($\pm$ 0.017) &\bfseries 0.0 ($\pm$ 0.0) &5.201 ($\pm$ 0.289) & $<$ 0.1 ($\pm$ 0.0)& 3.654 ($\pm$ 0.349)& 9.9 ($\pm$ 0.0) \\
    Trig7-20k   & \bfseries 0.530 ($\pm$ 0.036)&\bfseries 0.0 ($\pm$ 0.0) & 3.423 ($\pm$ 0.467) & $<$ 0.1 ($\pm$ 0.0)& 3.660 ($\pm$ 0.287)& 26.3 ($\pm$ 0.1) \\
    \bottomrule
\end{tabular}}
\end{table}

The percentages of invalid expressions generated by the approaches show that our approach always produces syntactically correct expressions, while GVAE and CVAE sometimes fail to produce valid outputs. The fraction of such expressions is quite small when the GVAE approach is used (see the explanation above) but quite significant when CVAE is used. Lastly, we can notice that, as expected, longer expressions are harder to recreate and thus have higher edit distance and a higher percentage of invalid expressions than shorter ones, provided enough training data is used.

\subsubsection{Training efficiency and the latent space dimensionality}\label{sec:eval-hvae-eff}
We proceed to test our conjectures about the efficiency of training the generators of mathematical expressions. We expect that \HVAE would require less training data and a lower dimensionality of the latent space to achieve the same levels of reconstruction error in comparison to other approaches. The latter is especially important because of the exploration of the latent space, which is more efficient in low-dimensional latent spaces.

\begin{figure*}[h]
     \centering
     \resizebox{\linewidth}{!}{
     \begin{subfigure}[b]{0.49\textwidth}
         \centering
         \includegraphics[width=\textwidth]{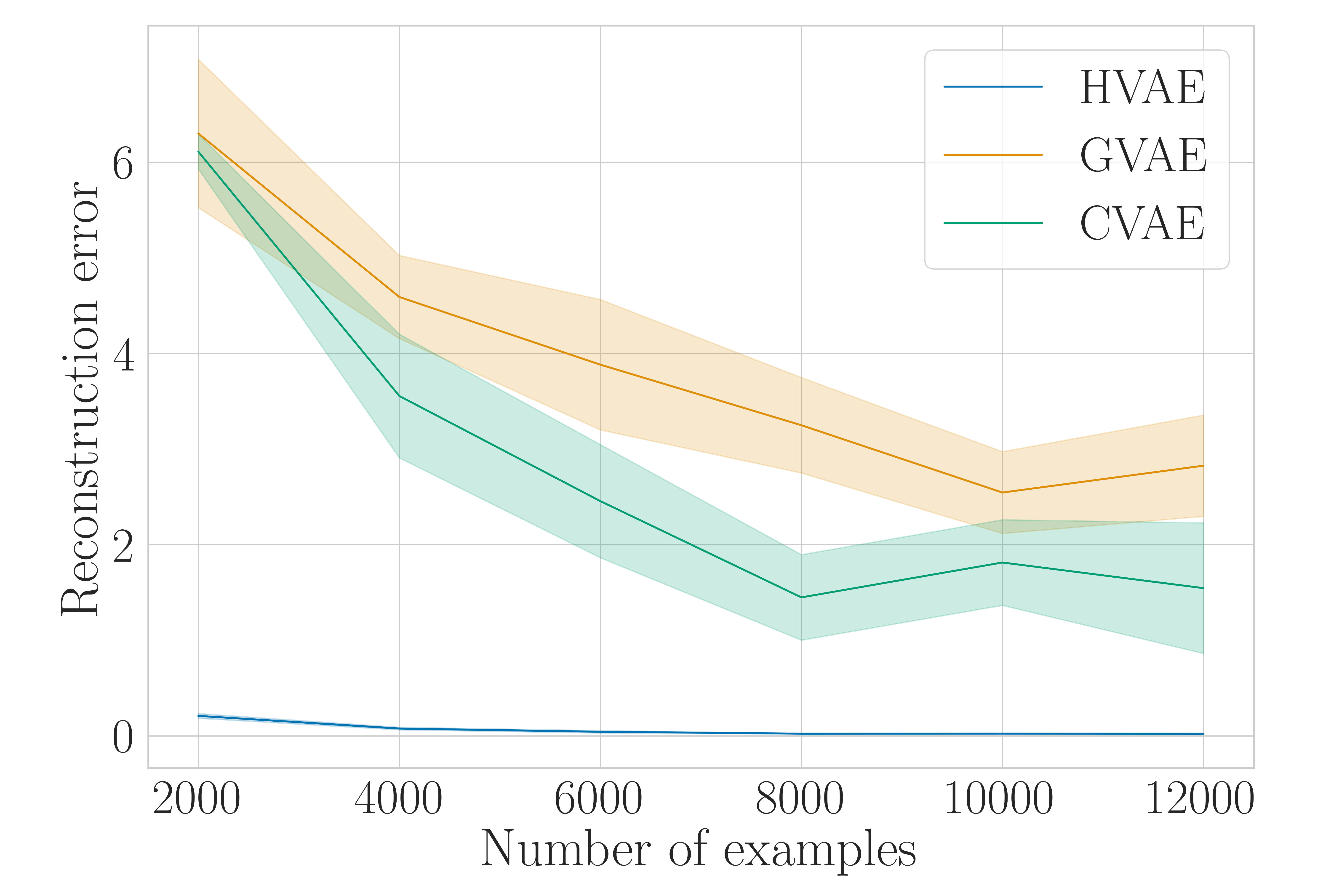}
         \caption{Size of the training set}
         \label{fig:examples}
     \end{subfigure}
     \hfill
     \begin{subfigure}[b]{0.49\textwidth}
         \centering
         \includegraphics[width=\textwidth]{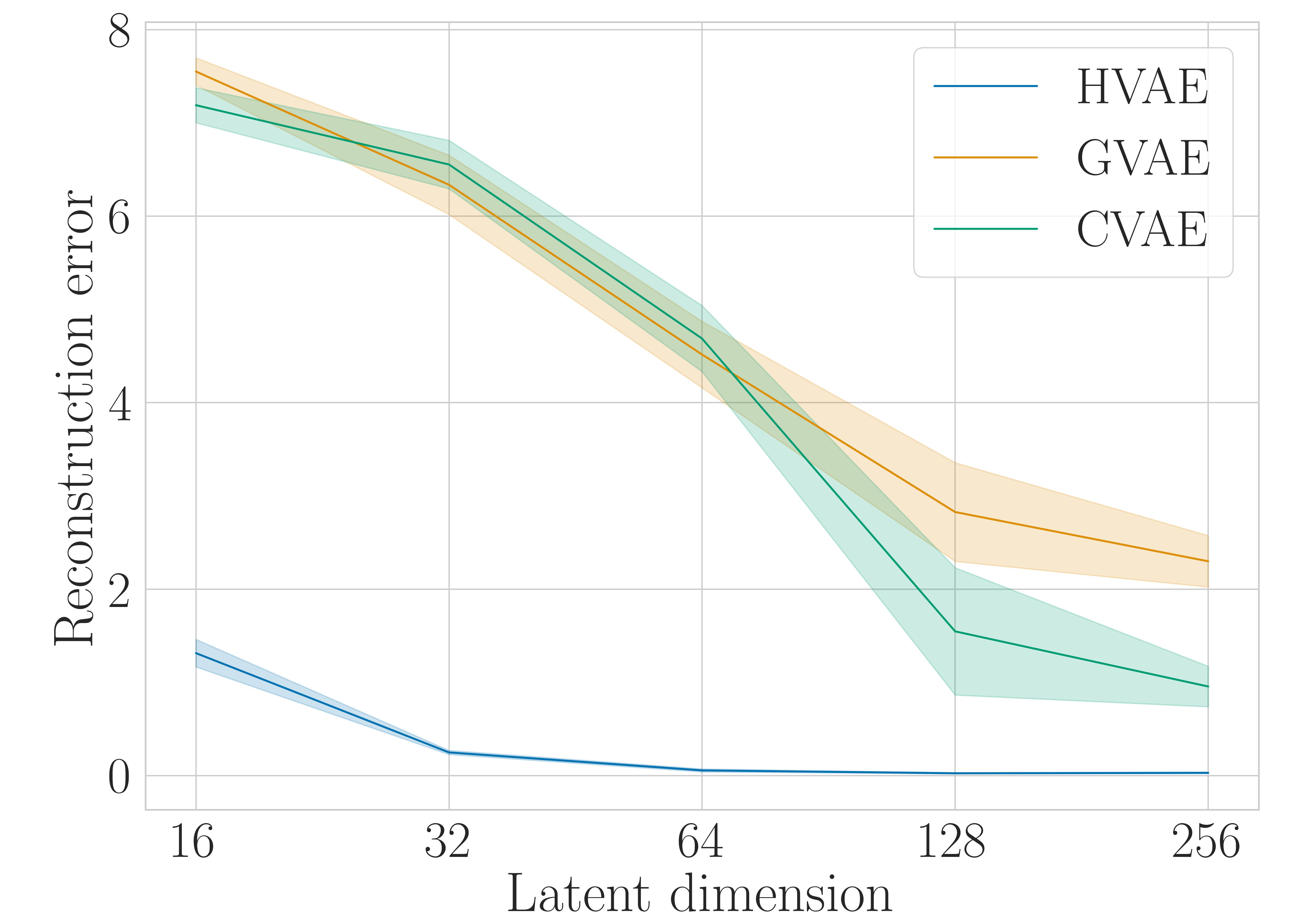}
         \caption{Dimensionality of the latent space}
         \label{fig:latent-dim}
     \end{subfigure}
    }
    \caption{The impact of the (a) training data set size and (b) dimensionality of the latent space on the reconstruction error of the three autoencoders.}
\end{figure*}

Figure~\ref{fig:examples} depicts the impact of the number of expressions in the training set on the reconstruction error for the three different generative models. Again, \HVAE significantly outperforms the other two VAEs. Its reconstruction error is estimated to be consistently lower than $0.25$, even when trained on $2$ thousand examples only. This error is an order of magnitude lower than the lowest error of $1.5$ achieved by the second best model, GVAE, when trained on the whole data set of $12$ thousand examples.

Figure~\ref{fig:latent-dim} shows the impact of the dimensionality of the latent space on the reconstruction error across different VAEs. In line with the previous results, \HVAE significantly outperforms both CVAE and GVAE. \HVAE with latent space of dimension $16$ performs on par or better than GVAE and CVAE with latent spaces of $256$ dimensions. We can see that the reconstruction error quickly raises when the latent space dimension is less than $32$, but otherwise, the reconstruction error is consistently low. Even with a latent space size of $16$, our approach is still comparable to the other two methods with a latent space of dimension $256$. This allows us to reduce the dimensionality of the latent space by two orders of magnitude, which makes \HVAE an excellent candidate for generating expressions for symbolic regression.

The reason for the superior efficiency of \HVAE is the use of expression trees, as subexpressions are always encoded into the same code, regardless of their position in the expression. This significantly reduces the space of possible codes and allows for training the model in a way that better generalizes to the repetitive subexpressions (subpatterns) it encounters.

\subsubsection{Latent space smoothness}\label{sec:eval-hvae-li}
Finally, we expect the latent space of \HVAE to be smooth in the sense that samples close to the latent representation of the input expression are decoded into expressions similar to the one given as input. We investigate the validity of this conjecture by applying linear interpolation (performing a \emph{homotopic} transformation) between two expressions in the latent space. Assume that we are given two expressions, $A$ and $B$. Using the model, we encode them into their latent representations $z_A$ and $z_B$. We can then generate new latent representations $z_\alpha$ by combining the two representations with the formula $z_\alpha = (1-\alpha)\cdot z_A + \alpha\cdot z_B$, where $\alpha \in \{i / n: i\in\mathbb{N} \land i \leq n\}$. Decoding the latent representations $z_\alpha$ in a smooth latent space should produce intermediate expressions that represent a smooth transition from $A$ to $B$ in $n$ steps.

\begin{table*}[htb!]
    \centering
    \caption{Examples of linear interpolation between two expressions in the latent spaces of the three VAEs. Expressions that are colored red are syntactically incorrect. Here, we set $n=4$ and $\alpha = i / 4, 0\leq i\leq 4$.}
    \resizebox{.8\textwidth}{!}{
	\begin{tabular}{cccc}
    \toprule
    $\alpha$ & HVAE & GVAE & CVAE \\\midrule
    Expression A & $c\cdot\cos c + \frac{c}{x} + \frac{x}{\sin x}$ & $c\cdot\cos c + \frac{c}{x} + \frac{x}{\sin c}$ & $c\cdot\cos c + \frac{c}{x}+\frac{x}{\sin c}$\\ 
    $\alpha$ = 0 & $c\cdot x + \frac{c}{x} + \frac{x}{\sin x}$ & $c\cdot\cos c + \frac{c}{x} + \frac{x}{\sin c}$ & $c\cdot\cos c + \frac{x}{x^x}\cdot\sin x$\\ 
    $\alpha$ = 0.25 & $c\cdot\sin c + \frac{c}{x} + x$ & $c + \frac{c}{\cos c} + c\cdot x$ & $\frac{c}{\cos c} + c - \frac{x}{c}\cdot\sin x$\\
    $\alpha$ = 0.5 & $\frac{c\cdot\sin c}{c} + x$ & $c + \frac{x}{c^{\cos c}}\cdot x$ & $c-\cos c - \frac{x}{x}\cdot\sin c$\\
    $\alpha$ = 0.75 & $\frac{\sin (x-\sin c)}{x}$ & $x + c\cdot\frac{c}{x}-x$ & $\color{red} \cos ( x(c)-c\cdot$ \\
    $\alpha$ = 1 & $\frac{\sin (x-c)}{x}$ & $\frac{\sin (x-c)}{x}$ & $\frac{\sin (x-c)}{x}$ \\ 
    Expression B & $\frac{\sin (x-c)}{x}$ & $\frac{\sin (x-c)}{x}$ & $\frac{\sin (x-c)}{x}$ \\\midrule
    Expression A & $x-\sin (c\cdot x)$ & $x-\sin (c\cdot x)$ & $x-\sin (c\cdot x)$\\ 
    $\alpha$ = 0 & $x-\sin (c\cdot x)$ & $\frac{x}{\sin (c\cdot x)}$ & $x-\sin (c\cdot x)$\\
    $\alpha$ = 0.25 & $x\cdot \sin c - \sin(c\cdot x)$ & $\frac{x}{\sin (c\cdot x)}$ & $x-\cos (c\cdot x)$\\
    $\alpha$ = 0.5 & $c\cdot \sin x +x$ & $c+\cos c$ & $ \color{red} \frac{c}{\cos}-\cos\cdot c)-c$\\
    $\alpha$ = 0.75 & $c\cdot \cos \frac{x}{c}+c$ & $c\cdot x\cdot\frac{\cos (\frac{x}{c})}{c}$ & $c\cdot x\cdot\frac{\cos (\frac{c}{x})}{c}$\\
    $\alpha$ = 1 & $c\cdot x\cdot \cos \frac{x}{c}+c$ & $c\cdot x\cdot\cos (\frac{x}{c})+c$ & $c\cdot x\cdot\cos (\frac{x}{c}) + c$\\ 
    Expression B & $c\cdot x\cdot \cos \frac{x}{c}+c$ & $c\cdot x\cdot\cos (\frac{x}{c})+c$ & $c\cdot x\cdot\cos (\frac{x}{c}) + c$\\\bottomrule
\end{tabular}
}
	\label{tab:li-all}
\end{table*}

Table~\ref{tab:li-all} shows the results of such a linear interpolation in the latent spaces of the different VAEs. \HVAE and GVAE produce continuous latent spaces where the transition from expression $A$ to expression $B$ is indeed smooth. CVAE also produces a smooth transition, but some of the intermediate expressions might be syntactically incorrect. The second interpolation in the lower part of the table is an example of a smooth transition in the \HVAE latent space. We can see that at each step only a few expression symbols change and that these changes are rarely redundant. Appendix~\ref{app:li} provides further examples of interpolations with visualizations of the expression trees.

\subsection{Evaluating \HVAESR}
In the second series of experiments, we evaluate the performance of \HVAESR. We start the section by introducing the experimental setup (Section~\ref{sec:eval-sr-setup}). We then report on the impact of the dimensionality of the latent space on the performance of symbolic regression (Section~\ref{sec:eval-sr-dim}). Furthermore, we compare the performance of \HVAESR with that of other methods for symbolic regression on the Nguyen benchmark (Section~\ref{sec:eval-sr-nguyen}) and report the performance of \HVAESR on the Feynman benchmark (Section~\ref{sec:eval-sr-feynman}).

\subsubsection{Experimental setup}\label{sec:eval-sr-setup}
\textbf{Data sets.} The Nguyen~\cite{Uy2011} benchmark contains eight equations with one non-target variable and two equations with two non-target variables. The right-hand sides of these equations are shown in the second column of Table~\ref{tab:dataset}. We generate two data sets (train and test) with five thousand simulated measurements for each equation. We use the train set to select the best expressions and the test set to evaluate their performance with the metrics described below. We sample points from the interval $[-20, 20]$ for equations 1-6, the interval $[0, 40]$ for equation 7, $[0, 80]$ for equation 8, and $[0, 20]$ for equations 9 and 10.

We further evaluate our approach on the $16$ equations involving up to two variables from the Feynman benchmark~\cite{Tegmark2020Feynman}. The right-hand sides of these equations are shown on in the last column of Table~\ref{tab:dataset}. Because equations in the Feynman benchmark represent real-world equations, each of the equations FM-3, FM-4, FM-5, and FM-7 contains two entries. Each entry comes with its own variables and data sets of measurements.

\begin{table}[h]
	\caption{Expressions from the Nguyen (first two columns on the left-hand side) and Feynman (last three columns on the right-hand side) benchmarks.}
	\label{tab:dataset}
    \centering
    \resizebox{.8\linewidth}{!}{
	\begin{tabular}{lrllr}
    \toprule
ID  & Expression                         & ID   & ID-Feynman                 & Expression \\ \midrule
NG-1  & $x^3 + x^2 + x$                    & FM-1  & I.6.2a             & $(2\pi)^{-0.5} e^{-\frac{x^2}{2}}$ \\  
NG-2  & $x^4 + x^3 + x^2 + x$              & FM-2  & I.6.2              & $(\sqrt{2\pi}\cdot y)^{-1} e^{-\frac{(x/y)^2}{2}}$  \\  
NG-3  & $x^5 + x^4 + x^3 + x^2 + x$        & FM-3  & I.12.1, I.12.5     & $xy$ \\  
NG-4  & $x^6 + x^5 + x^4 + x^3 + x^2 + x$  & FM-4  & I.14.4, II.8.31    & $0.5\, x y^2$  \\  
NG-5  & $\sin x^2 \cdot \cos x - 1$        & FM-5  & I.25.13, I.29.4    & $x/y$  \\  
NG-6  & $\sin x + \sin (x + x^2)$          & FM-6  & I.26.2             & $\arcsin (x\sin{y})$  \\ 
NG-7  & $\ln (x+1) + \ln (x^2 + 1)$        & FM-7  & I.34.27, III.12.43 & $(2\pi)^{-1}xy$   \\  
NG-8  & $\sqrt{x}$                         & FM-8  & I.39.1             & $ 1.5\, xy$\\   
NG-9  & $\sin x + \sin y^2$                & FM-9  & II.3.24            & $\frac{x}{4\pi y^2}$  \\  
NG-10 & $2\sin x \cdot \cos y$             & FM-10 & II.11.28           & $\frac{1+xy}{1-(0.\overline{3}\, xy)}$  \\  
       &                                   & FM-11 & II.27.18           & $xy^2$  \\               
       &                                   & FM-12 & II.38.14           & $\frac{x}{2\cdot(1 + y)}$ \\   
    \bottomrule
\end{tabular}}
\end{table}

\smallskip
\noindent
\textbf{Evaluation process.} We compare the performance of \HVAESR on the Nguyen benchmark equations to the performance of three other symbolic regression systems. ProGED~\cite{Brence2021ProGED} uses probabilistic grammars as generators of mathematical expressions. DSO~\cite{Petersen2021CharGen} combines deep neural networks with evolutionary algorithms. PySR~\cite{cranmer2023pysr} employs evolutionary optimization with operators directly applied to the expression trees. We run each system ten times on each equation and evaluate at most 100,000 sampled expressions. All approaches use the same library of tokens and/or grammars, further described in Appendix~\ref{app:grammars}. When running PySR, we allow fitting the values of the constant parameters since it can not be turned off in the implementation. The dimensionality of the latent space of \HVAE is set to 32; the ADAM optimizer uses the default learning rate. We elaborate on the setting of the dimensionality of the latent space in the next section. Appendix~\ref{app:nguyen} gives the complete report on the experiments in latent spaces with varying dimensions.

\smallskip
\noindent
\textbf{Estimating the performance.} We use three metrics: the number of successful reconstructions, i.e., runs leading to an equation equivalent to the original one; the mean $R^2$ of the best equation; and the number of expressions sampled to achieve reconstruction. We consider a run successful if we find an expression where the RMSE between the target and predicted values is lower than $10^{-10}$. To guarantee accurate reporting, we manually check if the original and expression with the lowest RMSE are equivalent. In each run, we use the expression with the lowest RMSE to calculate the bounded $R^2$ metric on the test set using the formula
\begin{equation}
    R^2(\hat{y}, y) = \max \left( 0, 1 - \frac{\sum_i (y_i - \hat{y}_i)^2}{\sum_i (y_i - \overline{y})^2}\right),
\end{equation}
where $\hat{y}_i$ denotes the predicted value of the target variable (calculated by using the equation), $y_i$ is the measured value of the target variable, and $\overline{y}$ is the mean value of $y$ in the training data set. Lastly, we show the average number (across the ten runs) of unique expressions considered before reconstructing the original equation. To this end, we count the unique expressions that the symbolic regression system has considered before the reconstructed equation is encountered in the generation process for all the methods that report this.

\subsubsection{The impact of the dimensionality of the latent space}\label{sec:eval-sr-dim}
Let us start with a series of computational experiments exploring the latent space for encoding mathematical expressions with random sampling. Here, we perform symbolic regression by taking randomly sampled points in the latent space and decoding them into expressions that are then evaluated on the measurements/data. The expression that fits the data best is selected as the candidate for the discovered equation.

\begin{table}[b]
  \caption{The performance of symbolic regression (number of successful reconstructions) by randomly sampling with CVAE, GVAE, and \HVAE on the Nguyen benchmark.}
  \label{tab:ng-random}
  \centering
  \resizebox{.6\linewidth}{!}{
    \begin{tabular}{lccccccccc}
    \toprule
     Equation\textbackslash Approach     & \multicolumn{3}{l}{CVAE}                                  & \multicolumn{3}{l}{GVAE}                                   & \multicolumn{3}{l}{HVAR}                                         \\
     Latent space size     & 32  & 64               & 128            & 32   & 64               & 128           & 32   & 64               & 128                 \\ \midrule
    NG-1     & 4           & 2            & 2          & 10            & 9            & 10          & 10           & 10          & 10           \\
    NG-2     & 0           & 0            & 0          & 2             & 4            & 3           & 10           & 5           & 9           \\
    NG-3     & 0           & 0            & 0          & 0             & 0            & 0           & 0            & 0           & 1           \\
    NG-4     & 0           & 0            & 0          & 0             & 1            & 0           & 0            & 0           & 0             \\
    NG-5     & 0           & 0            & 0          & 0             & 0            & 0           & 0            & 0           & 0            \\
    NG-6     & 0           & 0            & 0          & 2             & 0            & 0           & 4            & 4           & 0          \\
    NG-7     & 0           & 0            & 0          & 0             & 0            & 0           & 0            & 0           & 0           \\
    NG-8     & 10          & 10           & 3          & 10            & 10           & 10          & 10           & 10          & 10             \\
    \midrule
    Total/Mean  & 14     & 12         & 5         & 24           & 24        & 23    & 34          & 29 & 30 \\
    \bottomrule
\end{tabular}
  }
\end{table}

Table~\ref{tab:ng-random} shows the number of successful runs of the random sampling approaches based on the three VAEs, CVAE, GVAE, and \HVAE. In the further discussion of results, we use the name HVAR for HVAE with random sampling. We can see here a typical example of the curse of dimensionality at work. When the symbolic regression algorithm explores high-dimensional latent spaces, it can easily slip into parts of those spaces that do not lead to optimal equations. This shows that the ability of HVAE to encode mathematical expressions in low-dimensional latent spaces is crucial for the performance of symbolic regression with HVAR.

Based on the results of the experiments in Table~\ref{tab:ng-random} and Appendix~\ref{app:nguyen}, in the remainder of this section, we use 32-dimensional latent space for \HVAESR.

\subsubsection{Comparison on the Nguyen equations}\label{sec:eval-sr-nguyen}

\begin{table}[t]
  \caption{Comparison of the performance of symbolic regression (number of successful reconstructions, $R^2$, and number of evaluated equations) with random sampling on the Nguyen benchmark. We compare sampling from a manually-crafted probabilistic grammar (ProGED) with sampling using a trained \HVAE (HVAR).}
  \label{tab:ng-proged-hvae}
  \centering
  \resizebox{.7\linewidth}{!}{
    \begin{tabular}{lcccccc}
    \toprule
    & \multicolumn{3}{l}{ProGED~\cite{Brence2021ProGED}} & \multicolumn{3}{l}{HVAR (Ours)} \\
    Name      & Successful   & Average $R^2$                & Evaluated                 & Successful   & Average $R^2$               & Evaluated           \\ \midrule
    NG-1      & \bfseries 10 & \bfseries  1.00 ($\pm$ 0.00) & 2374 ($\pm$ 1451)         & \bfseries 10 & \bfseries 1.00 ($\pm$ 0.00)           & \bfseries 901 ($\pm$ 1332)     \\
    NG-2      & 2            & 1.00 ($\pm$ 0.01)            & \bfseries 7680 ($\pm$ 670)          & \bfseries 10 & \bfseries 1.00 ($\pm$ 0.00)           & 9729 ($\pm$ 5337)   \\
    NG-3      & 0            & 1.00 ($\pm$ 0.01)            & NA                        & 0            & 1.00 ($\pm$ 0.01)           & NA                   \\
    NG-4      & 0            & 1.00 ($\pm$ 0.01)            & NA                        & 0            & 1.00 ($\pm$ 0.01)           & NA                  \\
    NG-5      & 0            & 0.01 ($\pm$ 0.01)            & NA                        & 0            & 0.00 ($\pm$ 0.00)           & NA                  \\
    NG-6      & 0            & 0.60 ($\pm$ 0.11)            & NA                        & \bfseries 4  & \bfseries 0.81 ($\pm$ 0.20)           & \bfseries 37619 ($\pm$ 2773)  \\
    NG-7      & 0            & 0.99 ($\pm$ 0.01)            & NA                        & 0            & 0.99 ($\pm$ 0.01)           & NA                  \\
    NG-8      & \bfseries 10 & \bfseries 1.00 ($\pm$ 0.00)  & \bfseries 319 ($\pm$ 287) & \bfseries 10 & \bfseries 1.00 ($\pm$ 0.00)           & 392 ($\pm$ 456)      \\
    NG-9      & 1            & 0.56 ($\pm$ 0.14)            & \bfseries 12602 ($\pm$ 0)           & \bfseries 5            & \bfseries 0.83 ($\pm$ 0.21)           & 23236 ($\pm$ 11844)  \\
    NG-10     & 0            & 0.65 ($\pm$ 0.11)            & NA                        & 0            & 0.55 ($\pm$ 0.08)           & NA                  \\ \midrule
    Total/Mean & 23          & 0.78 ($\pm$ 0.31)            &                           & \bfseries 39 & \bfseries 0.81 ($\pm$ 0.31)           &                     \\
    \bottomrule
\end{tabular}
  }
\end{table}

In the next series of experiments, we compare the performance of HVAR, the random sampling method using \HVAE, to the one of ProGED---the latter samples mathematical expressions using manually crafted probabilistic grammar. Table~\ref{tab:ng-proged-hvae} reports the results of the comparison. The results show that the generator used within \HVAE is not worse than the probabilistic grammar. To our surprise, HVAR outperforms ProGED significantly. First, it successfully reconstructs five (of the ten) equations from the Nguyen benchmark in ten runs, one more than ProGED. Second, for the three equations of NG-2, NG-6, and NG-9, the reconstruction is achieved faster, i.e., by evaluating fewer candidate expressions. 

Furthermore, we check the contribution of the evolutionary approach in \HVAESR over the random sampling method HVAR. To this end, we compare the last three columns of Table~\ref{tab:ng-proged-hvae} with the last three columns of Table~\ref{tab:ng-hvae-dso}. \HVAESR successfully reconstructs all ten equations from the Nguyen benchmarks in at least one of the ten runs. In three cases, the equations are reconstructed in every run. Note also that the successful reconstructions with \HVAESR require fewer evaluations of candidate equations than the random sampling approaches.

\begin{table}[h]
  \caption{Comparison of the performance of the symbolic regression systems \HVAESR, DSO, and PySR on the Nguyen benchmark.}
  \label{tab:ng-hvae-dso}
  \centering
  \resizebox{.9\linewidth}{!}{
    \begin{tabular}{lcccccccc}
    \toprule
          & \multicolumn{3}{l}{EDHiE (our)}                                                         & \multicolumn{3}{l}{DSO~\cite{Petersen2021CharGen}}  & \multicolumn{2}{l}{PySR~\cite{cranmer2023pysr}} \\
    Name  & Successful           & Mean $R^2$                  & Evaluated                    & Successful   & Mean $R^2$                  & Evaluated                  & Successful & Mean $R^2$ \\ \midrule
    NG-1  & \bfseries 10         & \bfseries 1.00 ($\pm$ 0.00) & \bfseries 573 ($\pm$ 261)    & \bfseries 10 & \bfseries 1.00 ($\pm$ 0.00) & 4565 ($\pm$ 327)           & \bfseries 10 &  1.00  ($\pm$ 0.00) \\
    NG-2  & \bfseries 10         & \bfseries 1.00 ($\pm$ 0.00) & \bfseries 5803 ($\pm$ 4148)  & \bfseries 10 & \bfseries 1.00 ($\pm$ 0.00) & 12206 ($\pm$ 9186)         & \bfseries 10 &  1.00  ($\pm$ 0.00) \\
    NG-3  & 6                    & 1.00 ($\pm$ 0.01)           & 20931 ($\pm$ 4858)           & \bfseries 10 & \bfseries 1.00 ($\pm$ 0.00) & \bfseries 8053 ($\pm$ 3766)& 2 &   1.00  ($\pm$ 0.01) \\
    NG-4  & 3                    & 1.00 ($\pm$ 0.01)           & \bfseries 21346 ($\pm$ 4479) & \bfseries 8  & \bfseries 1.00 ($\pm$ 0.01) & 32946 ($\pm$ 15613)        & 0 &   0.99  ($\pm$ 0.01) \\
    NG-5  & \bfseries 3          & \bfseries 0.32 ($\pm$ 0.45) & \bfseries 20615 ($\pm$ 8394) & 0            & 0.00 ($\pm$ 0.00)           & NA                         & 0 &   0.16  ($\pm$ 0.15) \\
    NG-6  & \bfseries 8          & \bfseries 0.88 ($\pm$ 0.14) & \bfseries 12772 ($\pm$ 7923) & 1            & 0.59 ($\pm$ 0.15)           & 49599 ($\pm$ 0)            & 4 &   0.86  ($\pm$ 0.13) \\
    NG-7  & 8                    & 1.00 ($\pm$ 0.01)           & \bfseries 19203 ($\pm$ 3595) & \bfseries 10 & \bfseries 1.00 ($\pm$ 0.00) & 22579 ($\pm$ 10264)        & 7 &   0.99  ($\pm$ 0.01) \\
    NG-8  & \bfseries 10         & \bfseries 1.00 ($\pm$ 0.00) & \bfseries 405 ($\pm$ 174)    & \bfseries 10 & \bfseries 1.00 ($\pm$ 0.00) & 5521 ($\pm$ 1779)          & \bfseries 10 &  \bfseries 1.00  ($\pm$ 0.00) \\
    NG-9  & 8                    & 0.95 ($\pm$ 0.15)           & \bfseries 7041 ($\pm$ 3933)  & 2            & 0.60 ($\pm$ 0.20)           & 39786 ($\pm$ 28197)        & \bfseries 10 &  \bfseries 1.00  ($\pm$ 0.00) \\
    NG-10 & \bfseries 1          & 0.70 ($\pm$ 0.17) & \bfseries 31863 ($\pm$ 6970) & 0            & 0.56 ($\pm$ 0.10)           & NA                         & \bfseries 1 & \bfseries  0.80  ($\pm$ 0.16) \\
    \midrule
    Total/Mean & \bfseries 66 & \bfseries 0.89 ($\pm$ 0.21)& & 61 & 0.78 ($\pm$ 0.31)& & 54 &  0.88  ($\pm$ 0.26) \\
    \bottomrule
\end{tabular}
  }
\end{table}

Table~\ref{tab:ng-hvae-dso} compares \HVAESR with PySR, which uses evolutionary operators on expression trees directly (i.e., without embedding them into a latent space), and DSO, that similarly to our approach, combines deep learning with evolutionary optimization. Overall, \HVAESR performs better than the other two methods across all metrics\footnote{PySR does not report the number of evaluated equation. Hence, we could not include them in Table~\ref{tab:ng-hvae-dso}.}: it achieves the highest total number of successful reconstructions. \HVAESR has more successful reconstructions for five equations than PySR and less for a single equation, NG-9. The superior performance of \HVAESR relative to PySR indicates that evolutionary optimization is more efficient in the latent space than in the space of expression trees. For four equations, \HVAESR achieves successful reconstruction more often than DSO. In the two instances of reconstructing NG-3 and NG-4, DSO achieves success twice as often as our method.

Finally, note that the experiments on the Nguyen benchmark were performed on noise-free synthetic data. The results of the experiments on synthetic data with added noise, reported in Appendix~\ref{app:noise}, show that \HVAESR is robust to noise: The increasing noise level has little effect on the reconstruction success rate while significantly increasing the rank of the successfully reconstructed equation in the list of evaluated equations, sorted with respect to increasing RMSE. Appendix~\ref{app:nguyen} also includes additional results on the Nguyen benchmark by random sampling of CVAE, GVAE, and \HVAE latent space with varying number of dimensions.

\begin{table}[b]
  \caption{Results of \HVAESR on the 16 equations from the Feynman database that include at most two non-target variables.}
  \label{tab:feynman}
  \centering
  \resizebox{.5\linewidth}{!}{
    \begin{tabular}{lccc}
    \toprule
    Name  & Successful   & Mean $R^2$                   & Evaluated                                            \\ \midrule
    FM-1    & 10          & 1.00 ($\pm$ 0.00)            & 4311 ($\pm$ 1914)        \\
    FM-2    & 0           & 0.98 ($\pm$ 0.01)            & NA                       \\
    FM-3.1  & 10          & 1.00 ($\pm$ 0.00)            & 38 ($\pm$ 37)            \\
    FM-3.2  & 10          & 1.00 ($\pm$ 0.00)            & 53 ($\pm$ 28)            \\
    FM-4.1  & 10          & 1.00 ($\pm$ 0.00)            & 184 ($\pm$ 123)          \\
    FM-4.2  & 10          & 1.00 ($\pm$ 0.00)            & 188 ($\pm$ 204)          \\
    FM-5.1  & 10          & 1.00 ($\pm$ 0.00)            & 63 ($\pm$ 44)            \\
    FM-5.2  & 10          & 1.00 ($\pm$ 0.00)            & 101 ($\pm$ 109)          \\
    FM-6    & 0           & 0.99 ($\pm$ 0.01)            & NA                       \\
    FM-7.1  & 10          & 1.00 ($\pm$ 0.00)            & 43 ($\pm$ 36)            \\
    FM-7.2  & 10          & 1.00 ($\pm$ 0.00)            & 39 ($\pm$ 39)            \\
    FM-8    & 10          & 1.00 ($\pm$ 0.00)            & 62 ($\pm$ 45)            \\
    FM-9    & 10          & 1.00 ($\pm$ 0.00)            & 950 ($\pm$ 72)           \\
    FM-10   & 5           & 0.99 ($\pm$ 0.01)            & 22668 ($\pm$ 21676)      \\
    FM-11   & 10          & 1.00 ($\pm$ 0.00)            & 62 ($\pm$ 42)            \\
    FM-12   & 10          & 1.00 ($\pm$ 0.00)            & 924 ($\pm$ 795)          \\ \midrule
    Total/Mean  & 135     & 1.00 ($\pm$ 0.01)            &                            \\
    \bottomrule
\end{tabular}

  }
\end{table}

\subsubsection{Results on the Feynman equations}\label{sec:eval-sr-feynman}
In this section, we evaluate the ability of \HVAESR to reconstruct real equations from the domain of physics included in the Feynman database. Table~\ref{tab:feynman} presents the results of symbolic regression on a subset of 16 equations from the database with up to two non-target variables. \HVAESR successfully reconstructs $13$ equations in all the runs. Most of these equations are simple; thus, \HVAESR explores small search spaces comprising less than two hundred evaluated expressions. A more complex equation FM-10 is reconstructed in five out of ten runs exploring more than 20 thousand expressions on average. The equation FM-6 could not be reconstructed in any of the runs since it includes the function $\arcsin$ that has not been included in our token library. Finally, \HVAESR fails to reconstruct the most complex equation FM-2.

\section{Discussion and conclusion}
We introduce a novel variational autoencoder for hierarchies, \HVAE, that can be efficiently trained to generate valid mathematical expressions represented as expression trees. Compared to generators based on variational autoencoders for sequences, \HVAE has three significant advantages. First, it consistently generates valid expressions. Second, its performance is robust even for small training sets: \HVAE trained from only two thousand expressions achieves much lower reconstruction error than sequential VAEs trained from 12 thousand examples. Third, the \HVAE operating in $32$-dimensional latent space has a lower reconstruction error than sequential VAEs with comparable latent spaces.

The ability of \HVAE to encode mathematical expressions in a low-dimensional latent space makes it an excellent proxy for exploring the search space of candidate expressions in symbolic regression. Indeed, when performing a random search through the latent space, we achieve comparable performance with a random search through the space of candidate expressions defined by a manually crafted probabilistic grammar. \HVAESR, a symbolic regression system that performs evolutionary optimization in the latent space of the \HVAE, significantly outperforms methods based on random search and achieves performance comparable to the state-of-the-art symbolic regression system DSO based on a similar combination of evolutionary algorithms and deep learning. The comparison of \HVAESR with PySR, a genetic programming approach operating on expression trees directly, shows the benefit of performing evolutionary optimization in the latent space.

\HVAE has been used here for symbolic regression, but its potential to efficiently generate and encode hierarchies makes it useful in many different contexts, e.g., generating molecular structures or more general symbolic expressions. Analysis of its performance in these application domains is a promising direction for further research. Moreover, the ability of \HVAE to learn from small corpora of expressions might prove helpful in retraining the generator after each generation of the evolutionary search, much like the iterative learning in DSO. This will narrow its focus to generating better expressions, leading to more accurate equations. In general, training the generator on expressions involved in mathematical models that have proved useful in a domain of interest will enable seamless integration and transfer of background knowledge in symbolic regression.

\bmhead{Code availability}
The implementation of \HVAE and \HVAESR and the scripts needed for performing their evaluation, presented in this article, can be found at \url{https://github.com/smeznar/HVAE}.

\bmhead{Acknowledgments}
The authors acknowledge the financial support of the Slovenian Research Agency via the research core funding No.~P2-0103, project No.~N2-0128, and by the ARRS Grant for young researchers (first author). The authors especially appreciate the helpful comments and suggestions by Nikola Simidjievski and the fruitful discussions within the SHED discussion group (with Jure Brence, Boštjan Gec, and Nina Omejc).

\section*{Declarations}
\begin{itemize}
\item Funding: The authors acknowledge the financial support of the Slovenian Research Agency via the research core funding No.~P2-0103, project No.~N2-0128, and by the ARRS Grant for young researchers (first author).
\item Conflict of interest/Competing interests: Not applicable
\item Ethics approval: Not applicable
\item Consent to participate: Not applicable
\item Consent for publication: Not applicable
\item Availability of data and materials: All data used in this work is available in repositories \url{https://github.com/smeznar/HVAE}.
\item Code availability: The code used in experiments is available in repositories \url{https://github.com/smeznar/HVAE}.
\item Authors' contributions: Conceptualization: Sebastian Me\v{z}nar, Ljup\v{c}o Todorovski; Methodology: Sebastian Me\v{z}nar, Ljup\v{c}o Todorovski; Writing - original draft preparation: Sebastian Me\v{z}nar, Ljup\v{c}o Todorovski; Writing - review and editing: Sebastian Me\v{z}nar, Ljup\v{c}o Todorovski, Sa\v{s}o D\v{z}eroski; Funding acquisition: Sa\v{s}o D\v{z}eroski; Resources: Sa\v{s}o D\v{z}eroski; Supervision: Ljup\v{c}o Todorovski, Sa\v{s}o D\v{z}eroski; Software: Sebastian Me\v{z}nar; Visualization: Sebastian Me\v{z}nar; Data curation: Sebastian Me\v{z}nar; Investigation: Sebastian Me\v{z}nar; Validation: Sebastian Me\v{z}nar.
\end{itemize}

\bibliography{bibliography}

\backmatter

\begin{appendices}

\section{Grammars and token libraries}
\label{app:grammars}
In the empirical evaluation of the hierarchical autoencoder, we use several context-free grammars and different token libraries. Grammars used to generate synthetic data sets are probabilistic. Mathematical expressions in the data sets with a name prefix of \textsl{AE} include the five common binary arithmetic operators and are generated using the following grammar:
\begin{align*}
    S &\to S A F \, [0.4] \; \vert \; F \, [0.6] \\
    A &\to + \, [0.5] \; \vert \; - \, [0.5] \\
    F &\to F B T \, [0.4] \; \vert \; T \, [0.6] \\
    B &\to \cdot \, [0.5] \; \vert \; / \, [0.5] \\    
    T &\to ( S ) \, [0.25] \; \vert \; c \, [0.375] \, \vert \, x \, [0.375]
\end{align*}

Data sets with a name prefix \textsl{Trig} include, in addition, the trigonometric functions of sine and cosine and are generated using the same grammar as the one above, with different productions for the non-terminal (syntactic category) $T$ and a new non-terminal $L$:
\begin{align*}
    T &\to ( S ) \, [0.15] \;\vert\; \cos (S) \, [0.05] \;\vert\; \sin (S) \, [0.05] \;\vert\; L \, [0.75] \\
    L &\to c \, [0.5] \;\vert\; x\, [0.5]. 
\end{align*}  

While we do not explicitly have the power operator in the grammars to be used during the generation of the data sets, exponentiation (and thus the power operator) can occur during the simplification of the generated expressions. Because of this, expressions in the data sets also contain the power operator.

In addition, GVAE also needs grammars for generating valid expressions. When applied to the data sets with a name prefix of \textsl{AE}, GVAE uses the following grammar:
\begin{align*}
    S &\to S + T \,\vert\,  S - T \,\vert\, S \cdot T \,\vert\, \frac{S}{T} \,\vert\, S ^ T \,\vert\, T \\
    T &\to (S) \,\vert\, x \,\vert\, c
\end{align*}

For the data sets with the name prefix of \textsl{Trig}, GVAE uses the same grammar as above with different productions for the non-terminal symbol $T$:
$$ T \to (S) \,\vert\, \sin ( S ) \,\vert\, \cos (S) \,\vert\, x \,\vert\, c. $$

CVAE uses the token library $\{+,-,\cdot, /, \hat{\mkern6mu}, x, c, (, ), \text{''} \}$ for data sets with the name prefix \textsl{AE}, and $\{+,-,\cdot, /, \hat{\mkern6mu}, \sin, \cos, x, c, (, ), \text{''} \}$ for data sets with the name prefix \textsl{Trig}.

For experiments on the Nguyen benchmark, we use the grammar:

\begin{align*}
    E &\to E + F \, [0.2] \; \vert \; E - F \, [0.2] \; \vert \; F \, [0.6] \\
    F &\to E \cdot T \, [0.2] \; \vert \; E / T \, [0.2] \; \vert \; T \, [0.6] \\
    T &\to V \, [0.4] \; \vert \; ( E ) P \, [0.2] \; \vert \; ( E ) \, [0.2] \; \vert \; R ( E ) \, [0.2] \\
    V &\to x \, [1.0] \\
    P &\to \hat{\mkern6mu}^2 \, [0.39] \; \vert \; \hat{\mkern6mu}^3 \, [0.26] \; \vert \; \hat{\mkern6mu}^4 \, [0.19] \; \vert \; \hat{\mkern6mu}^5 \, [0.16] \\
    R &\to \sin \, [0.2] \; \vert \; \cos \, [0.2] \; \vert \; e\hat{\mkern6mu} \, [0.2] \; \vert \; \log \, [0.2]  \; \vert \; \text{sqrt} \, [0.2] \\
\end{align*}

for ProGED, GVAE, and to generate training examples for \HVAE. \HVAE and DSO use the token library $\{x, +, -, \cdot, /, \hat{\mkern6mu}^2, \hat{\mkern6mu}^3, \hat{\mkern6mu}^4, \hat{\mkern6mu}^5, \sin, \cos, \exp, \log, \text{sqrt}\}$, while CVAE uses token $\{ (, ), \text{''}\}$ in addition to the tokens used by \HVAE and DSO. For expressions with two non-target variables, we add token $y$ and change the non-terminal symbol $V$ to:
\begin{align*}
    V &\to x \, [0.5] \; \vert \; y \, [0.5]. \\
\end{align*}

For experiments on the Feynman benchmark, we use the grammar:

\begin{align*}
    E &\to E + F \, [0.2] \; \vert \; E - F \, [0.2] \; \vert \; F \, [0.6] \\
    F &\to E \cdot T \, [0.2] \; \vert \; E / T \, [0.2] \; \vert \; T \, [0.6] \\
    T &\to V \, [0.4] \; \vert \; c \, [0.3] \; \vert \; A \, [0.3] \\
    A &\to ( E ) P \, [0.1] \; \vert \; ( E ) \, [0.55] \; \vert \; R ( E ) \, [0.35] \\
    V &\to x \, [1.0] \\
    P &\to \hat{\mkern6mu}^2 \, [0.8] \; \vert \; \hat{\mkern6mu}^3 \, [0.2] \\
    R &\to \sin \, [0.25] \; \vert \; \cos \, [0.25] \; \vert \; e\hat{\mkern6mu} \, [0.25]  \; \vert \; \text{sqrt} \, [0.25] \\
\end{align*}
to generate training examples for \HVAE and tokens $\{x, c, +, -, \cdot, /, \hat{\mkern6mu}^2, \hat{\mkern6mu}^3, \sin, \cos, \exp, \text{sqrt}\}$. For expressions with two non-target variables, we add token $y$ and change the non-terminal symbol $V$ to:
\begin{align*}
    V &\to x \, [0.5] \; \vert \; y \, [0.5]. \\
\end{align*}

\section{Additional latent space smoothness results}\label{app:li}
Additional examples of linear interpolation with \HVAE are shown in Table~\ref{tab:li}. We can see that the space is continuous, as the expressions smoothly transition from Expression 1 to Expression 2. This is best seen in example 1 from the Trig5-15k data set, where in each step, only a few (relevant) symbols change. In the first step, $\sin c$ and $x$ change to $c$. Then in the next step, $c+c$ and $\sin x$ change to $x$. In the next step $\frac{x}{c}$ changes to $c$ and $x/c$ changes to $x\cdot\sin c$. In the last step $x+c$ changes to $x^c$.

\begin{table*}[htb!]
    \centering
    \resizebox{\textwidth}{!}{
	\begin{tabular}{cccc}
    \toprule
    Dataset & $\alpha$ & Example 1 & Example 2 \\\midrule
    &Expression A & $c \cdot (x+c) + \frac{x^c}{c}$ & $c-x\cdot c + x$ \\ 
    &$\alpha$ = 0 &$c \cdot (x+c) + \frac{x^c}{c}$ & $c-x\cdot c + x$ \\ 
    &$\alpha$ = 0.25 & $c \cdot (x+c) + \frac{x^c}{c}$ & $c-x\cdot c + x$ \\
    AE4-2k&$\alpha$ = 0.5 &$c\cdot x + \frac{x+c}{c}$ &  $c-c\cdot x + x$ \\
    &$\alpha$ = 0.75 &$c-x + \frac{x+c}{c}$ &  $\frac{c+c\cdot x}{c}$ \\
    &$\alpha$ = 1 &$c-x + \frac{x+c}{c}$ & $\frac{c+c\cdot x}{c}$ \\ 
    &Expression B &$c-x + \frac{x+c}{c}$ & $\frac{c+c\cdot x}{c}$ \\\midrule
    &Expression A & $\frac{c\cdot x^c + x^c}{c}+c$ & $\frac{c\cdot x}{c-x}+c-x+c$ \\ 
    &$\alpha$ = 0 & $\frac{c\cdot x^c + x^c}{c}+c$ & $\frac{c\cdot x}{c-x}+c-x+c$ \\ 
    &$\alpha$ = 0.25 & $\frac{c+x^c}{c}+c$ & $c\cdot (c-x)+c-x+c$ \\
    AE7-20k&$\alpha$ = 0.5 & $\frac{c\cdot x^c}{c}+\frac{x}{c}$ & $c\cdot x\cdot c-x+c$ \\
    &$\alpha$ = 0.75 & $c-\frac{x^c}{c}$ & $\frac{c\cdot x^c}{c+x}+x$ \\
    &$\alpha$ = 1 & $c-\frac{x^c}{c}$ & $\frac{c\cdot x^c}{c+x}+x$ \\ 
    &Expression B & $c-\frac{x^c}{c}$ & $\frac{c\cdot x^c}{c+x}+x$ \\\midrule
    &Expression A & $c+\sin c+\frac{\sin x}{x}-\frac{x}{c}$ & $x^c\cdot\cos c + x$ \\ 
    &$\alpha$ = 0 & $c+\sin c+\frac{\sin x}{x}-\frac{x}{c}$ & $x^c\cdot\cos c + x$ \\
    &$\alpha$ = 0.25 & $c+c+\frac{\sin x}{c}-\frac{x}{c}$ & $x\cdot\cos c + x$ \\
    Trig5-15k&$\alpha$ = 0.5 & $x+\frac{x}{c}-\frac{x}{c}$ & $c\cdot\cos c+\cos x^c$ \\
    &$\alpha$ = 0.75 & $x+c-x\cdot \sin c$ & $c+\cos\frac{x^c}{c}$ \\
    &$\alpha$ = 1 & $x^c-x\cdot \sin c$ & $c+\cos\frac{x^c}{c}$  \\ 
    &Expression B & $x^c-x\cdot \sin c$ & $c+\cos\frac{x^c}{c}$ \\\bottomrule
\end{tabular}
}
	\caption{Linear interpolation of examples in the \HVAE latent space. The first row shows examples from the AE4-2k data set, the second examples from the AE7-20k data set, and the third from the Trig5-15k data set. Here $n=4$ and $\alpha=\frac{i}{4}, 0\leq i\leq 4$.}
	\label{tab:li}
\end{table*}

While most of the time, expressions change gradually, this is not always the case. This can be best seen in the examples from the AE4-2k data set, where only the expressions at $\alpha=0.5$ differ from the starting ones.

\begin{figure*}[h!]
  \centering
  \includegraphics[width=\linewidth]{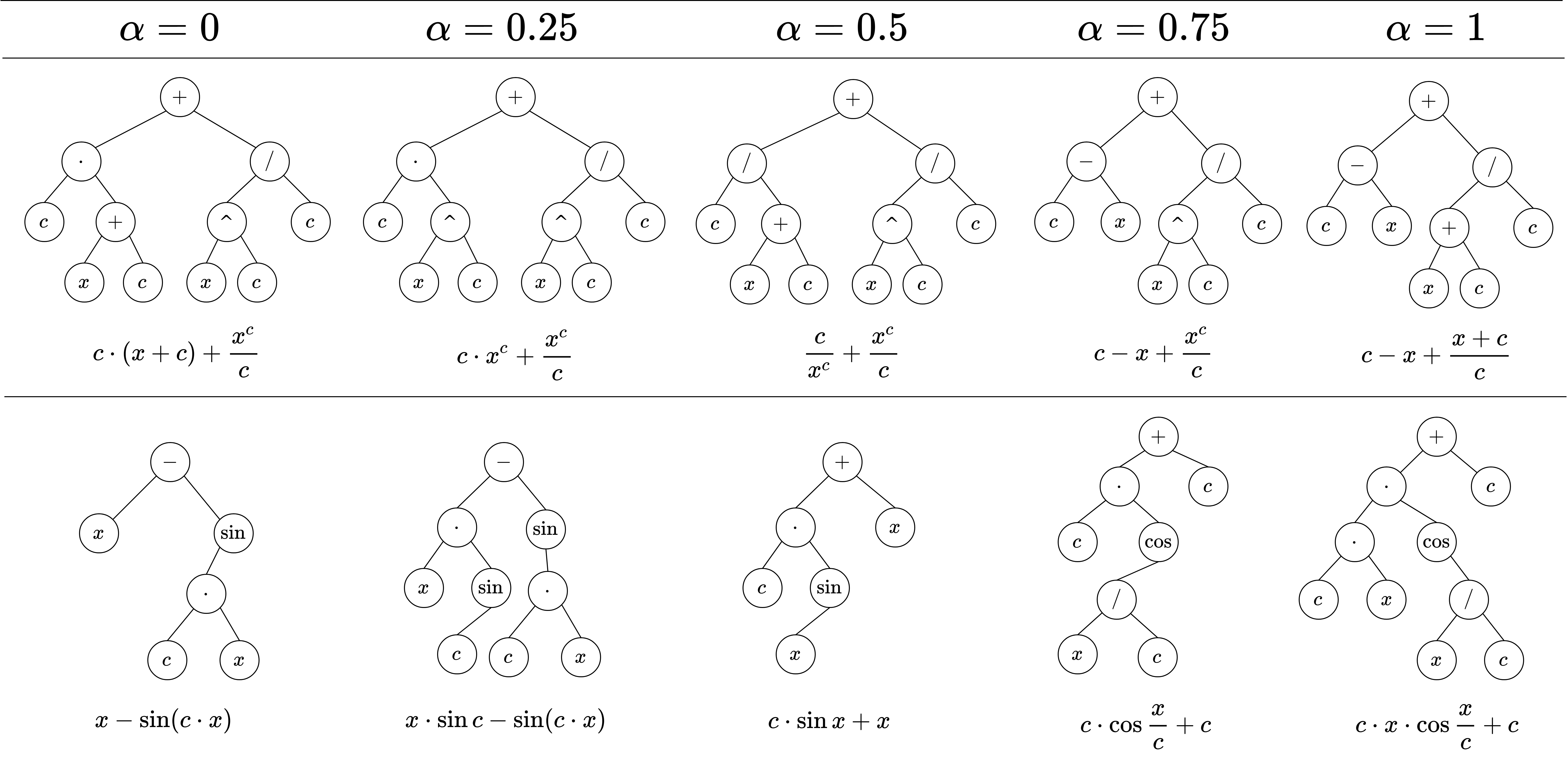}
  \caption{Examples of linearly interpolated mathematical expressions visualized as expression trees.}
  \label{fig:interpolation-trees}
\end{figure*}

Since we write expressions as a sequence, it is not always obvious how the underlying expression tree changes during interpolation. Visualization of the gradual change for two pairs of expressions is shown in Figure~\ref{fig:interpolation-trees}. 

Trees are being transformed by changing their structure and the symbols inside nodes. These transformations are interrelated as changing a constant or a variable into an operator also transforms the structure. This is best seen in the transition between $\alpha=0$ and $\alpha=0.25$ for the expression trees at the bottom of Figure~\ref{fig:interpolation-trees}. However, the structure of the expression tree does not always change. This is most noticeable for transitions between $\alpha=0$ and $\alpha=0.5$ for expression trees at the top of Figure~\ref{fig:interpolation-trees}. Here only an operator changes at each step. 

\section{Additional results on the Nguyen benchmark}\label{app:nguyen}
In this appendix, we show additional results of the empirical evaluation. First, we show the performance of HVAR on the first eight equations from the Nguyen benchmark with random sampling (By sampling points from the standardized Gaussian distribution) and different dimensions of the latent space in Section~\ref{app:hvar}. Next, we show the learning curves of approaches HVAR, ProGED, and \HVAESR in Section~\ref{app:lc}. After this, we show the performance of \HVAESR on noisy data in Section~\ref{app:noise}. Finally, we conclude this section in~\ref{app:other} by showing the performance of CVAE with random sampling and GVAE with both random sampling and an evolutionary algorithm. 

\subsection{Dimensions}\label{app:hvar}
Table~\ref{tab:sr-hvae} shows the results of evaluation using \HVAE with different dimensions of the latent vector space. Here new expressions are generated by randomly sampling points from the standardized Gaussian distribution and decoding them. We can see overall, models, where the dimension of the latent vector space is either $16$ or $32$, perform the best. \textsc{\HVAE 32} produces expressions that usually have a slightly higher mean $R^2$, while \textsc{\HVAE 16} usually needs to evaluate less unique expressions to generate the desired one. Because we prefer the number of successful runs and the mean $R^2$ metrics, we select the model with the latent vector space dimension $32$ for experiments where the \HVAE model is coupled together with evolutionary algorithms. 

\begin{table}[h]
  \caption{The performance of HVAR (number of successful reconstructions, $R^2$, and number of evaluated equations) with varying number of latent dimensions on the Nguyen benchmark.}
  \label{tab:sr-hvae}
  \centering
  \resizebox{\linewidth}{!}{
    \begin{tabular}{lcccccccccccc}
    \toprule
          & \multicolumn{3}{l}{HVAR 16}                                             & \multicolumn{3}{l}{HVAR 32}                                   & \multicolumn{3}{l}{HVAR 64}                          & \multicolumn{3}{l}{HVAR 128}\\
    Name  & Successful   & Mean $R^2$                   & Evaluated                 & Successful   & Mean $R^2$               & Evaluated           & Successful & Mean $R^2$        & Evaluated           & Successful & Mean $R^2$                   & Evaluated                           \\ \midrule
    NG-1  & 10           & 1.00 ($\pm$ 0.00)            & 500 ($\pm$ 378)           & 10           & 1.00 ($\pm$ 0.00)        & 901 ($\pm$ 1332)    & 10         & 1.00 ($\pm$ 0.00) & 1985 ($\pm$ 2270)   & 10         & 1.00 ($\pm$ 0.00)            & 1544 ($\pm$ 1523)           \\
    NG-2  & 10           & 1.00 ($\pm$ 0.00)            & 4407 ($\pm$ 2904)         & 10           & 1.00 ($\pm$ 0.00)        & 9729 ($\pm$ 5337)   & 5          & 1.00 ($\pm$ 0.01) & 14435 ($\pm$ 8279)  & 9          & 1.00 ($\pm$ 0.01)            & 14261 ($\pm$ 13074)         \\
    NG-3  & 1            & 1.00 ($\pm$ 0.01)            & 16595 ($\pm$ 0)           & 0            & 1.00 ($\pm$ 0.01)        & NA                  & 0          & 1.00 ($\pm$ 0.01) & NA                  & 1          & 1.00 ($\pm$ 0.01)            & 24562 ($\pm$ 0)             \\
    NG-4  & 0            & 1.00 ($\pm$ 0.01)            & NA                        & 0            & 1.00 ($\pm$ 0.01)        & NA                  & 0          & 1.00 ($\pm$ 0.01) & NA                  & 0          & 1.00 ($\pm$ 0.01)            & NA                          \\
    NG-5  & 0            & 0.00 ($\pm$ 0.00)            & NA                        & 0            & 0.00 ($\pm$ 0.01)        & NA                  & 0          & 0.01 ($\pm$ 0.01) & NA                  & 0          & 0.04 ($\pm$ 0.07)            & NA                          \\
    NG-6  & 3            & 0.75 ($\pm$ 0.18)            & 13345 ($\pm$ 425)         & 4            & 0.81 ($\pm$ 0.13)        & 37619 ($\pm$ 2773)  & 4          & 0.74 ($\pm$ 0.22) & 26169 ($\pm$ 12187) & 0          & 0.57 ($\pm$ 0.09)            & NA                          \\
    NG-7  & 0            & 1.00 ($\pm$ 0.01)            & NA                        & 0            & 0.99 ($\pm$ 0.01)        & NA                  & 0          & 1.00 ($\pm$ 0.01) & NA                  & 0          & 1.00 ($\pm$ 0.01)            & NA                          \\
    NG-8  & 10           & 1.00 ($\pm$ 0.00)            & 345 ($\pm$ 276)           & 10           & 1.00 ($\pm$ 0.00)        & 392 ($\pm$ 456)     & 10         & 1.00 ($\pm$ 0.00) & 233 ($\pm$ 171)     & 10         & 1.00 ($\pm$ 0.00)            & 273 ($\pm$ 205)             \\ \midrule
    Total/Mean  & \bfseries 34     & 0.84 ($\pm$ 0.33)            &                           & \bfseries 34           &\bfseries 0.85 ($\pm$ 0.32)        &                     & 29         & 0.84 ($\pm$ 0.33) &                     & 30         & 0.83 ($\pm$ 0.33)            &                             \\
    \bottomrule
\end{tabular}

  }
\end{table}

\subsection{Learning curves}\label{app:lc}
Optimization algorithms continually enhance solutions over time by iteratively exploring the input space to minimize an objective function or maximize performance. The learning curve serves as a valuable measure for evaluating algorithmic performance, illustrating how the chosen metric evolves as optimization progresses. Steep improvements in the learning curve indicate rapid convergence towards better solutions, while plateaus or slow convergence suggest challenges in finding superior solutions. By analyzing the learning curve, we can be gain an insight into the algorithm's effectiveness, convergence, stability, and potential for further improvement.

\begin{figure*}
     \centering
     \begin{subfigure}[b]{0.49\textwidth}
         \centering
         \includegraphics[width=\textwidth]{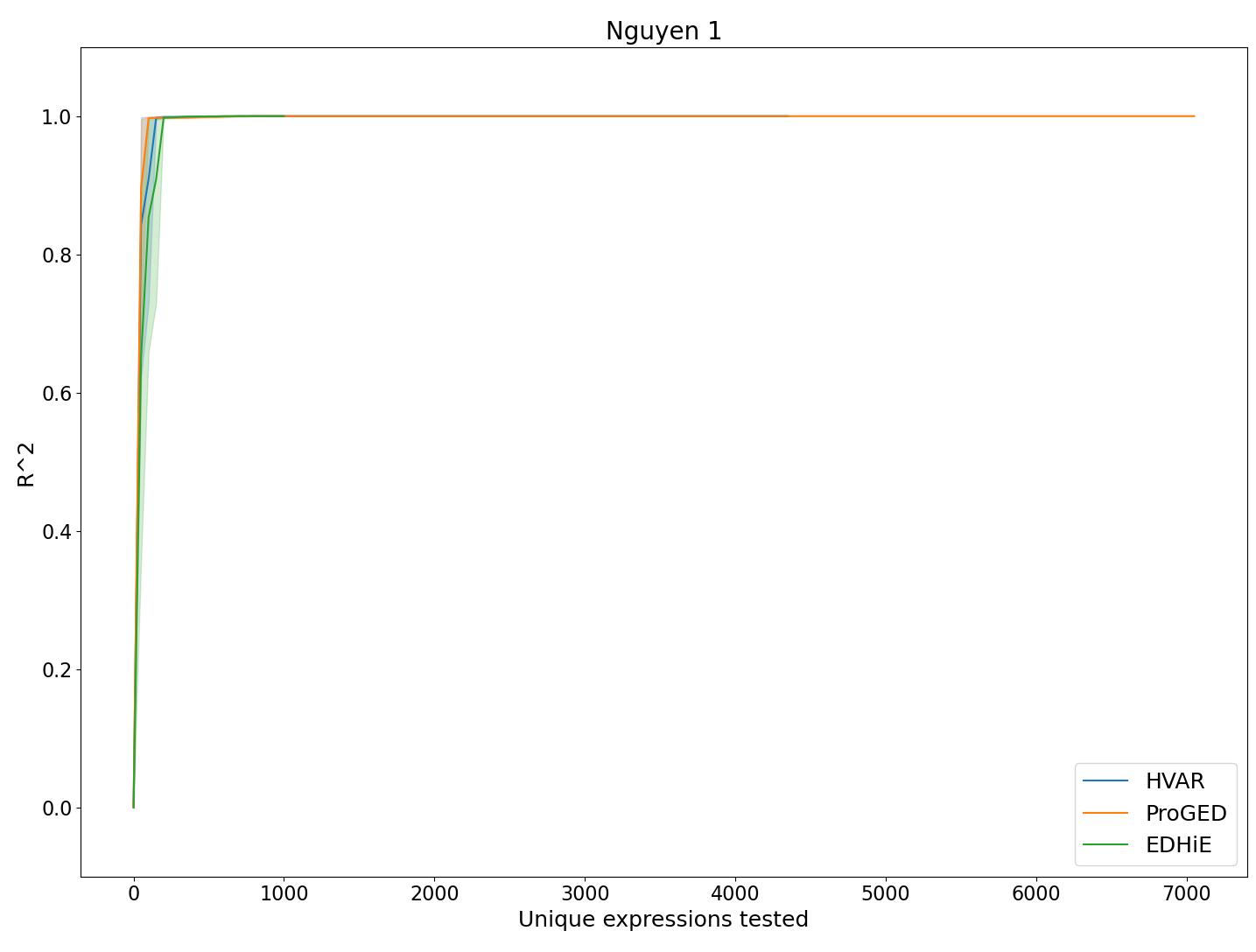}
         \caption{NG-1}
         \label{fig:lc1}
     \end{subfigure}
     \hfill
     \begin{subfigure}[b]{0.49\textwidth}
         \centering
         \includegraphics[width=\textwidth]{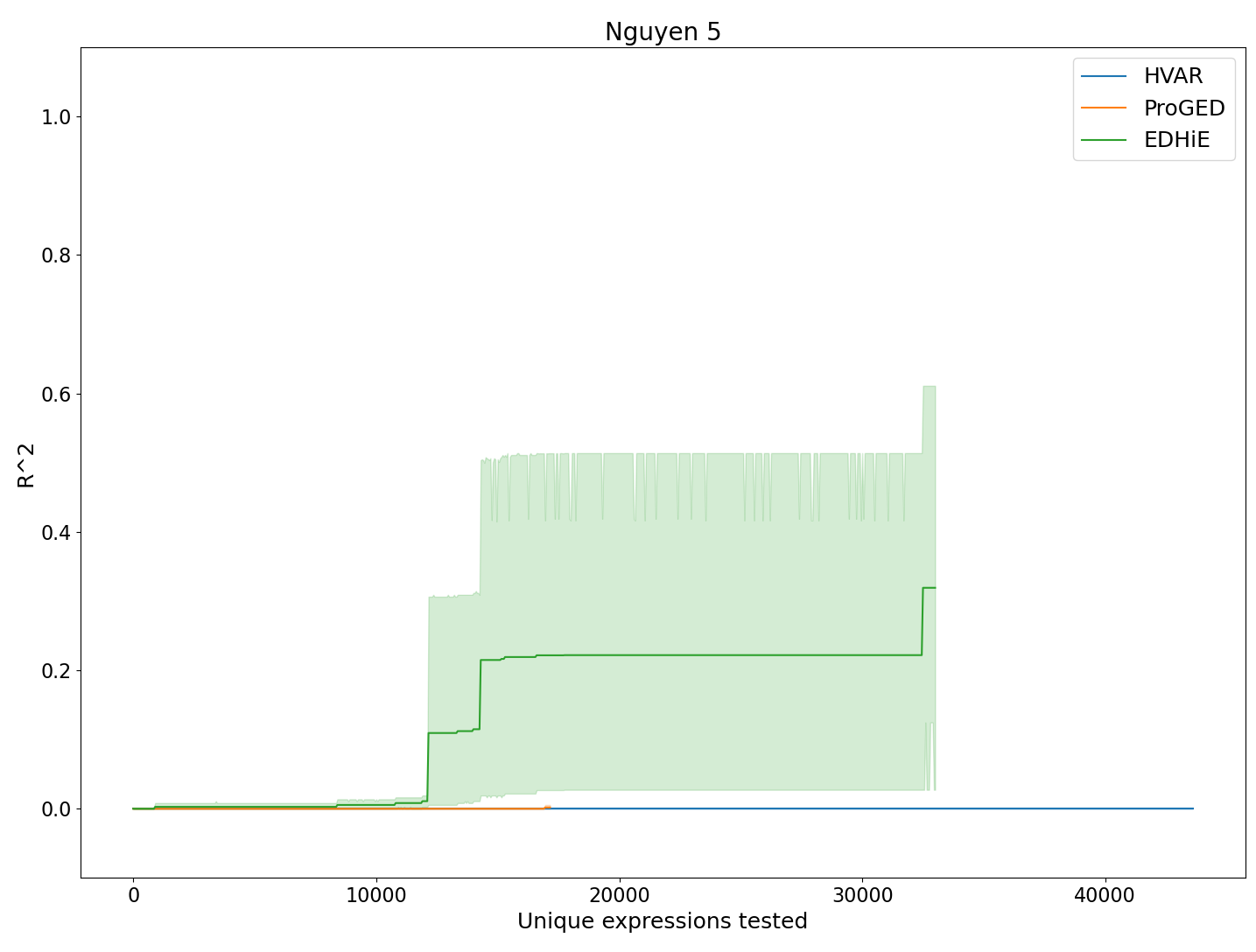}
         \caption{NG-5}
         \label{fig:lc5}
     \end{subfigure}

     \begin{subfigure}[b]{0.49\textwidth}
         \centering
         \includegraphics[width=\textwidth]{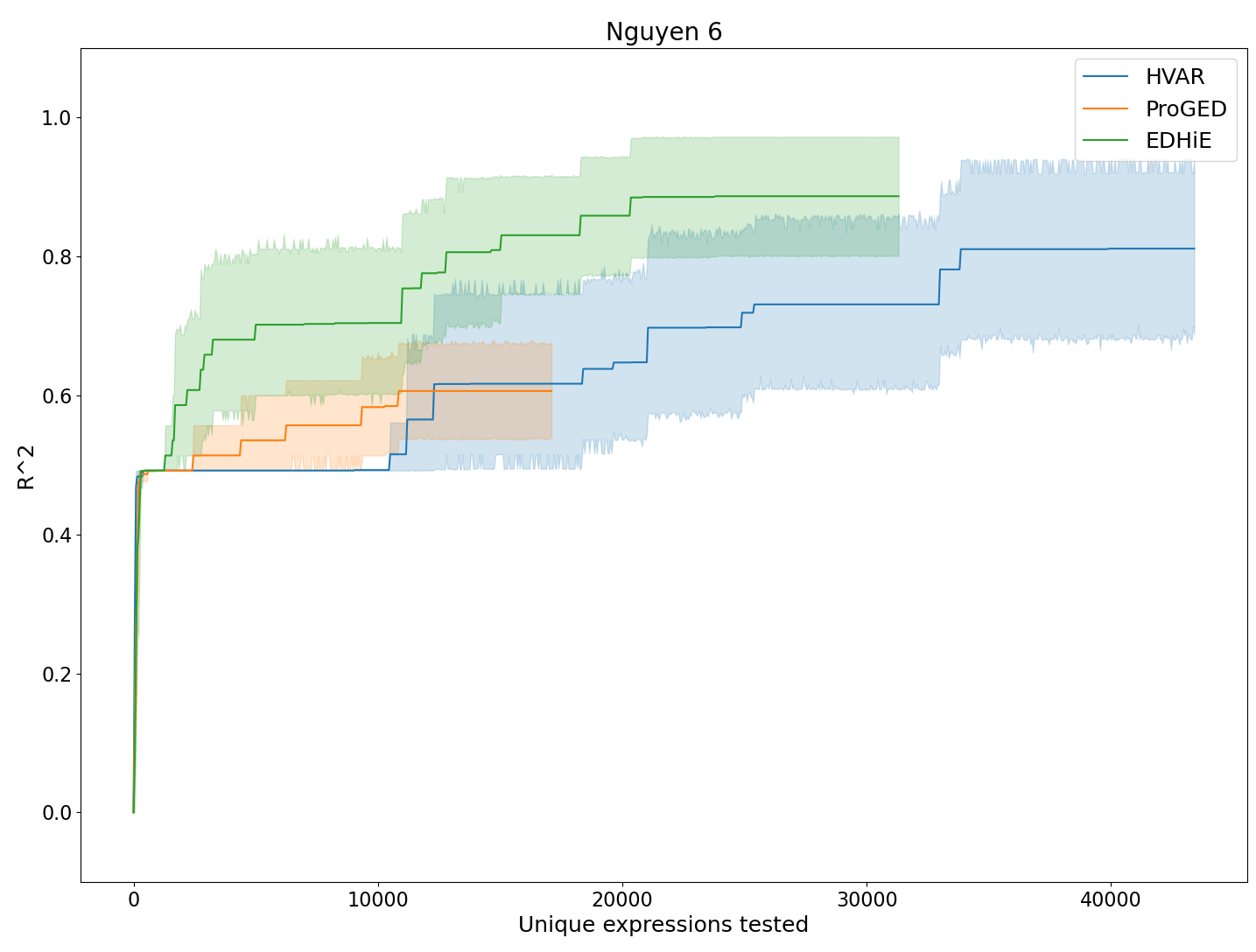}
         \caption{NG-6}
         \label{fig:lc6}
     \end{subfigure}
     \hfill
     \begin{subfigure}[b]{0.49\textwidth}
         \centering
         \includegraphics[width=\textwidth]{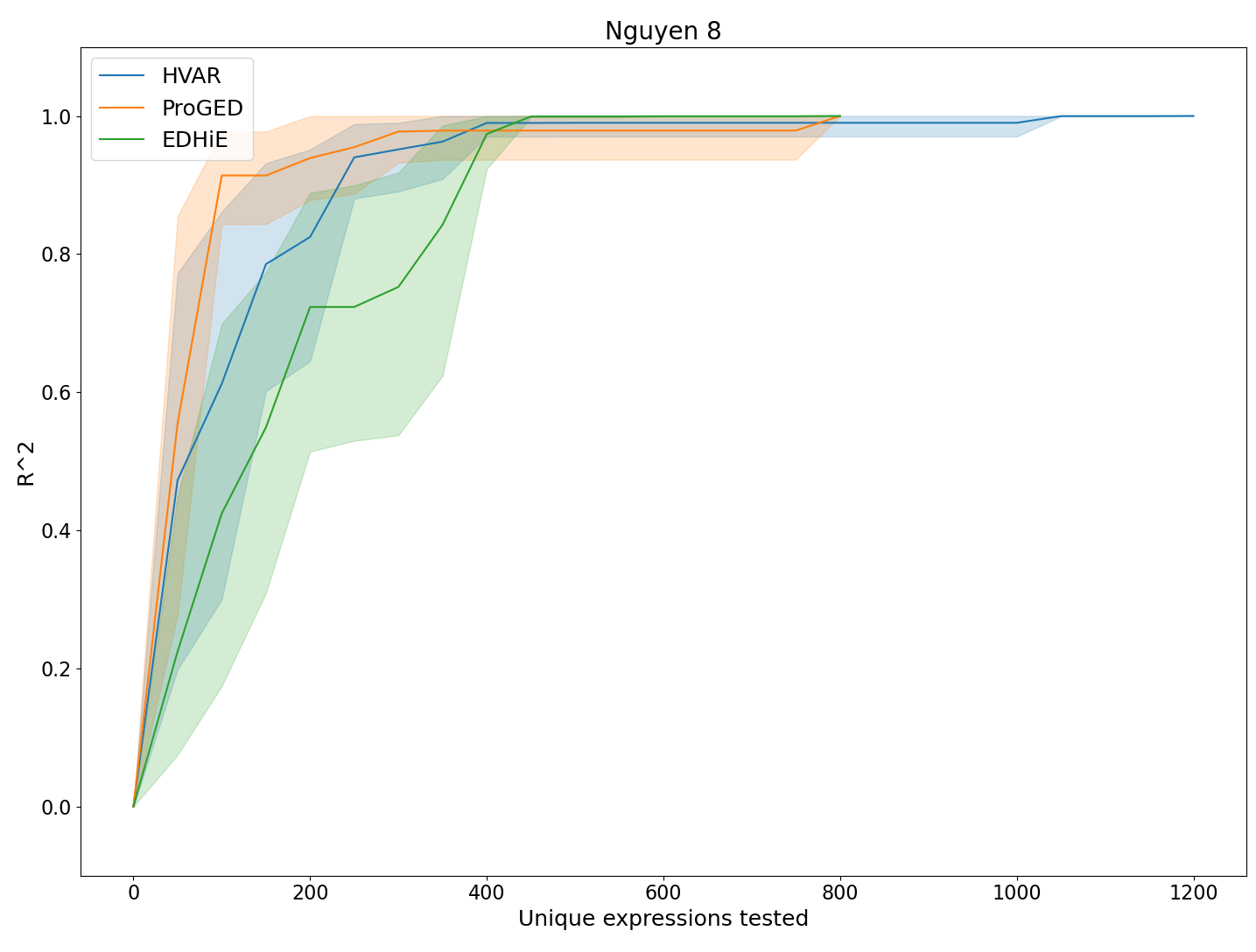}
         \caption{NG-8}
         \label{fig:lc8}
     \end{subfigure}

    \begin{subfigure}[b]{0.49\textwidth}
         \centering
         \includegraphics[width=\textwidth]{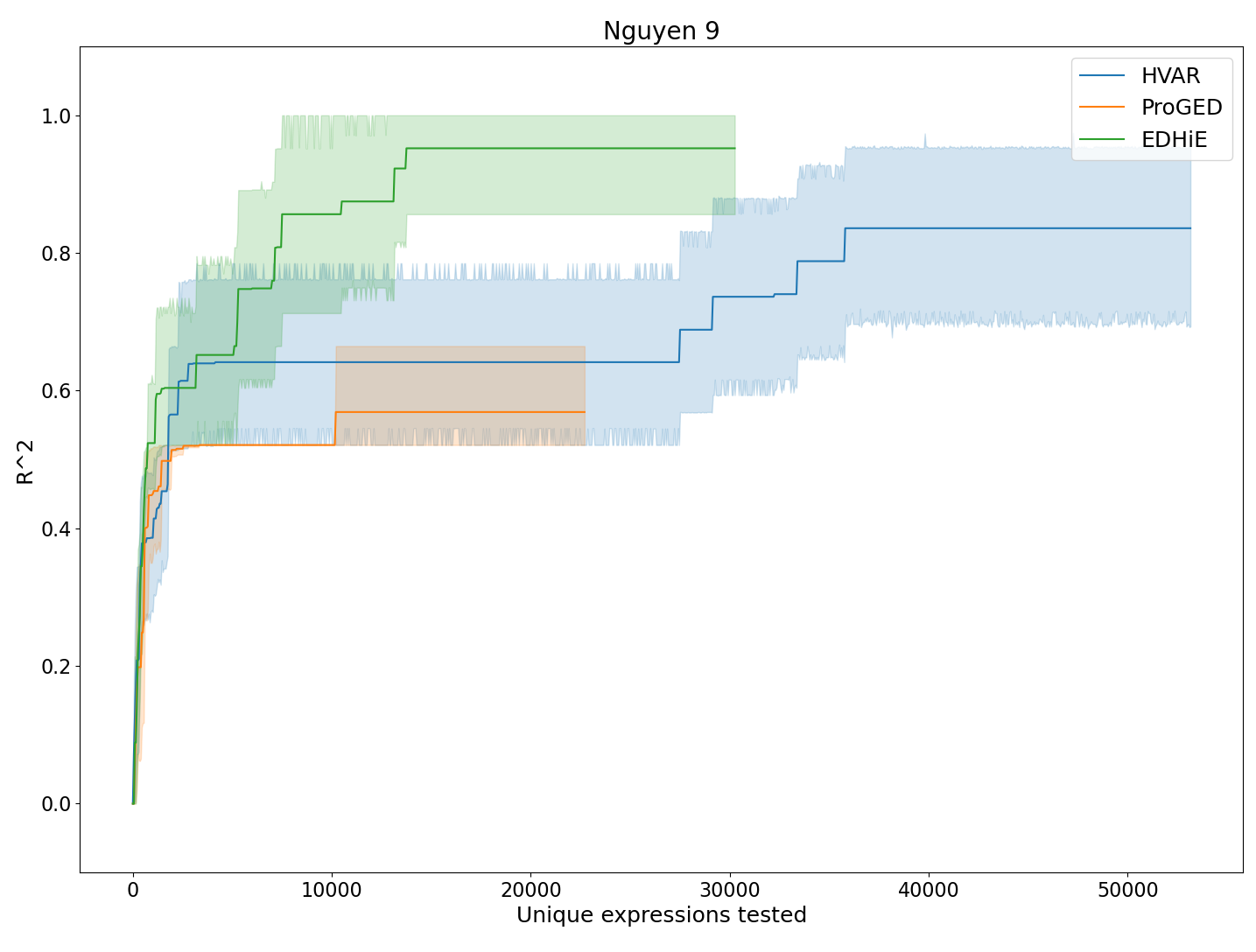}
         \caption{NG-9}
         \label{fig:lc9}
     \end{subfigure}
     \hfill
     \begin{subfigure}[b]{0.49\textwidth}
         \centering
         \includegraphics[width=\textwidth]{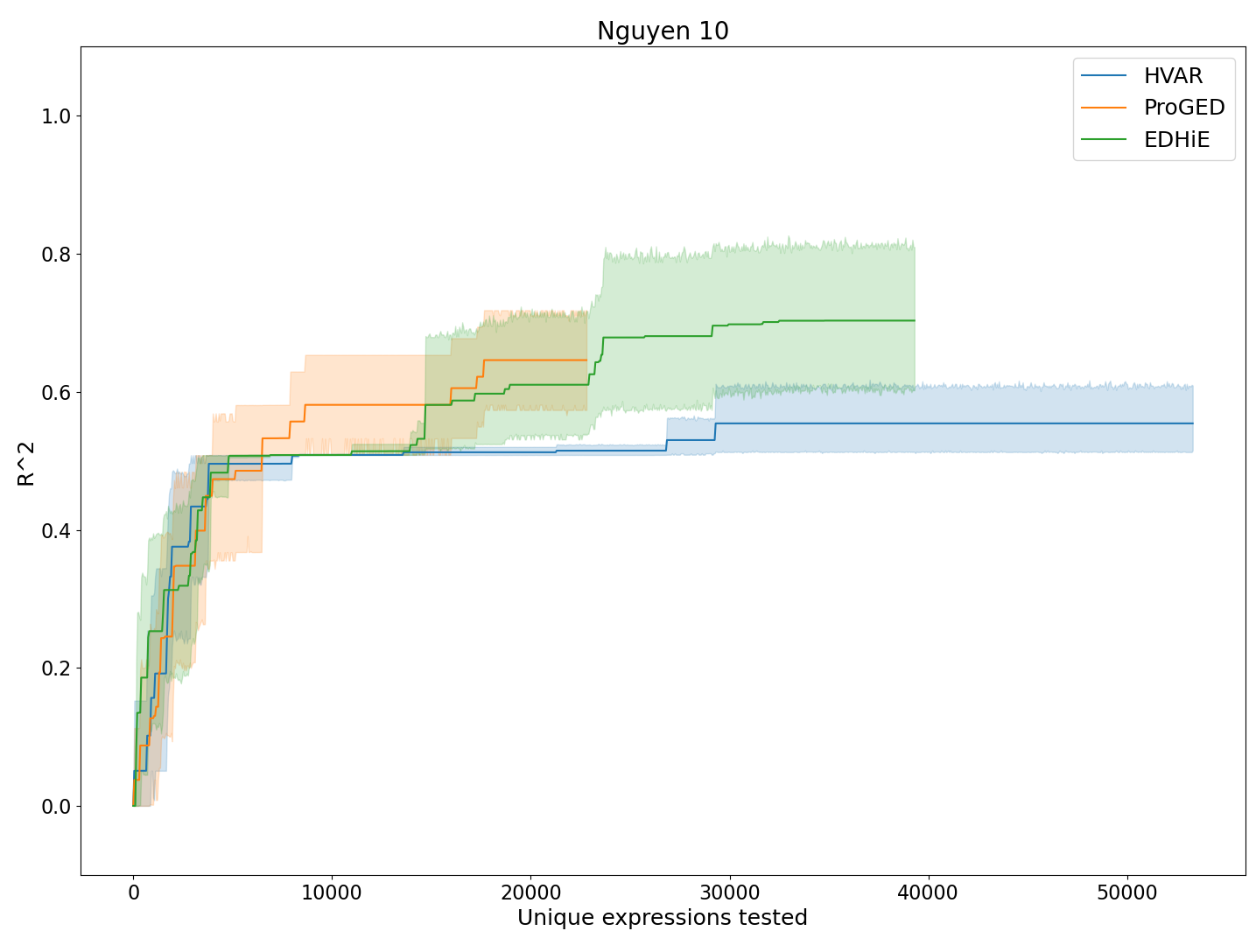}
         \caption{NG-10}
         \label{fig:lc10}
     \end{subfigure}
    \caption{Learning curves for HVAR, ProGED, and \HVAESR on the selected equations from the Nguyen benchmark. Curves for the equations \textit{NG-2}, \textit{NG-3}, \textit{NG-4}, and \textit{NG-7} are omitted since they resemble the curve for the \textit{NG-1} equation.\label{fig:curves}}
\end{figure*}

Figure~\ref{fig:curves} shows the learning curves of HVAR, ProGED, and \HVAESR on six equations from the Nguyen benchmark. We can see that overall \HVAESR performs the best as it achieves the highest $R^2$ score and needs to test the least expression to do so. This is not true on equation NG-8, where the other two approaches find the desired expressions quicker. This happens because all approaches find the desired expression quickly and the evolution part of \HVAESR does not yet come into effect. 

\subsection{Robustness to noise}\label{app:noise}
In practical scenarios, working with noisy data is common, making it crucial for symbolic regression approaches to perform well in the presence of noise. The performance of \HVAESR on noisy data is demonstrated in Table~\ref{tab:noise}. To generate noisy data sets, we sample values $\epsilon$ from a Gaussian distribution $\mathcal{N}(0, I)$ and add them to the target values $y$ using the formula $\tilde{y} = y\cdot (1 + \eta \epsilon)$, where $\eta$ represents the noise level.

To evaluate our approach on noisy data, we employ two metrics: the number of successful runs and the mean rank. We execute our approach on noisy data, rank all the generated expressions based on their RMSE, and evaluate these expressions using noiseless data. A run is considered successful if we find an expression that achieves an RMSE below $10^{-10}$ on noiseless data. In the case of a successful run, we record the rank of the first expression with an RMSE below $10^{-10}$ and use it to calculate the mean rank.

\begin{table}[h!]
  \caption{The performance of \HVAESR with varying level of noise added to synthetic data 
from the Nguyen benchmark.}
  \label{tab:noise}
  \centering
  \resizebox{\linewidth}{!}{
    \begin{tabular}{lcccccccccc}
    \toprule
    $\eta$ & 0.01 & & 0.02 & & 0.05 & & 0.1 & & 0.2 & \\
    Name & Successful & Mean Rank & Successful & Mean Rank & Successful & Mean Rank & Successful & Mean Rank & Successful & Mean Rank \\ \midrule
    NG-1 & 10 & 7.30 (6.74) 	& 10 & 6.00 (6.20) 	& 10 & 9.00 (8.04) 	& 10 & 14.60 (16.27) 	& 10 & 23.80 (13.91) 	\\
    NG-2 & 10 & 1.10 (0.30) 	& 10 & 1.20 (0.40) 	& 10 & 6.80 (3.09) 	& 10 & 175.10 (42.15) 	& 10 & 222.80 (28.72) 	\\
    NG-3 & 5 & 39.40 (4.41) 	& 3 & 32.67 (20.98) & 3 & 34.33 (9.81) 	& 2 & 50.50 (1.50) 	& 3 & 169.33 (25.85) 	\\
    NG-4 & 2 & 85.00 (38.00) 	& 4 & 57.00 (6.16) 	& 2 & 52.50 (6.50) 	& 0 &      NA      	& 2 & 210.50 (4.50) 	\\
    NG-5 & 4 & 1.00 (0.00) 	& 1 & 1.00 (0.00) 	& 2 & 1.00 (0.00) 	& 4 & 1.00 (0.00) 	& 5 & 1.00 (0.00) 	\\
    NG-6 & 6 & 1.00 (0.00) 	& 6 & 1.00 (0.00) 	& 6 & 1.00 (0.00) 	& 9 & 1.00 (0.00) 	& 7 & 1.00 (0.00) 	\\
    NG-7 & 10 & 1.00 (0.00) 	& 8 & 1.00 (0.00) 	& 9 & 1.00 (0.00) 	& 9 & 1.11 (0.31) 	& 8 & 10.00 (3.46) 	\\
    NG-8 & 10 & 1.00 (0.00) 	& 10 & 1.50 (1.50) 	& 10 & 1.70 (1.19) 	& 10 & 3.30 (3.44) 	& 10 & 5.10 (4.39) 	\\
    NG-9 & 9 & 1.00 (0.00) 	& 7 & 1.00 (0.00) 	& 9 & 1.00 (0.00) 	& 8 & 1.00 (0.00) 	& 8 & 1.00 (0.00) 	\\
    NG-10 & 1 & 1.00 (0.00) 	& 0 &      NA      	& 1 & 1.00 (0.00) 	& 1 & 1.00 (0.00) 	& 1 & 1.00 (0.00) 	\\ \midrule
    Total & 67 & 7.33 (18.38) & 59 & 7.37 (16.10) & 62 & 6.61 (11.91) & 63 & 32.75 (65.10) & 64 & 55.42 (87.83) \\
    \bottomrule
\end{tabular}
  }
\end{table}

The results demonstrate that the number of successful runs remains relatively consistent across different noise levelsm indicating the robustness of our approach. Additionally, we can see that the mean rank increases as the amount of noise rises. This outcome is expected, as higher noise levels allow more expressions to overfit the noisy data. In practical applications, expressions that overfit can be eliminated by assigning complexity scores to each expression and selecting less complex expressions from the Pareto front.

\subsection{Performance of CVAE and GVAE on symbolic regression}\label{app:other}
Table~\ref{tab:sr-cvae} shows the results of the CVAE baseline. Models presented in the table use the same parameters as they do in Section~\ref{sec:eval-hvae-setup} apart from the dimension of the latent vector space. We can see that this baseline performs very poorly, as it finds only the two simplest equations. The main reason for this is the high number of invalid expressions (more than $96.8\%$) the baseline produces.

\begin{table}[h]
  \caption{Results of symbolic regression by random sampling of the CVAE latent space with varying number of dimensions on the Nguyen benchark.}
  \label{tab:sr-cvae}
  \centering
  \resizebox{\linewidth}{!}{
    \begin{tabular}{lccccccccc}
    \toprule
          & \multicolumn{3}{l}{CVAE 32}                                             & \multicolumn{3}{l}{CVAE 64}                                   & \multicolumn{3}{l}{CVAE 128}                                 \\
    Name  & Successful   & Mean $R^2$                   & Invalid                   & Successful   & Mean $R^2$               & Invalid             & Successful & Mean $R^2$        & Invalid                     \\ \midrule
    NG-1  & 4            & 1.00 ($\pm$ 0.01)            & 96851 ($\pm$ 57)          & 2            & 1.00 ($\pm$ 0.01)        & 98131 ($\pm$ 57)    & 2          & 1.00 ($\pm$ 0.01) & 99243 ($\pm$ 25)           \\
    NG-2  & 0            & 1.00 ($\pm$ 0.01)            & 96816 ($\pm$ 38)          & 0            & 1.00 ($\pm$ 0.01)        & 98143 ($\pm$ 38)    & 0          & 1.00 ($\pm$ 0.01) & 99247 ($\pm$ 18)          \\
    NG-3  & 0            & 0.99 ($\pm$ 0.01)            & 96857 ($\pm$ 59)          & 0            & 0.99 ($\pm$ 0.01)        & 98131 ($\pm$ 47)    & 0          & 0.97 ($\pm$ 0.07) & 99250 ($\pm$ 22)          \\
    NG-4  & 0            & 0.99 ($\pm$ 0.01)            & 96867 ($\pm$ 39)          & 0            & 0.99 ($\pm$ 0.01)        & 98151 ($\pm$ 60)    & 0          & 0.99 ($\pm$ 0.01) & 99238 ($\pm$ 17)            \\
    NG-5  & 0            & 0.00 ($\pm$ 0.00)            & 96846 ($\pm$ 46)          & 0            & 0.00 ($\pm$ 0.00)        & 98148 ($\pm$ 46)    & 0          & 0.00 ($\pm$ 0.00) & 99239 ($\pm$ 18)           \\
    NG-6  & 0            & 0.49 ($\pm$ 0.01)            & 96843 ($\pm$ 55)          & 0            & 0.49 ($\pm$ 0.02)        & 98127 ($\pm$ 43)    & 0          & 0.09 ($\pm$ 0.15) & 99227 ($\pm$ 24)         \\
    NG-7  & 0            & 0.71 ($\pm$ 0.29)            & 96837 ($\pm$ 49)          & 0            & 0.60 ($\pm$ 0.32)        & 98108 ($\pm$ 27)    & 0          & 0.14 ($\pm$ 0.28) & 99249 ($\pm$ 31)          \\
    NG-8  & 10           & 1.00 ($\pm$ 0.00)            & 96801 ($\pm$ 48)          & 10           & 1.00 ($\pm$ 0.00)        & 98146 ($\pm$ 39)    & 3          & 0.58 ($\pm$ 0.42) & 99229 ($\pm$ 34)             \\
    \midrule
    Total/Mean  & \bfseries 14     & \bfseries 0.77 ($\pm$ 0.34)            & \bfseries 96839 ($\pm$ 20)          & 12           & 0.75 ($\pm$ 0.34)        & 98135 ($\pm$ 13)    & 5          & 0.59 ($\pm$ 0.42) & 99240 ($\pm$ 8) \\
    \bottomrule
\end{tabular}
  }
\end{table}

Lastly, Table~\ref{tab:sr-gvae} shows the results of the GVAE baseline. We can see that GVAE performs better than CVAE but worse than \HVAE. For the \textsc{GVAE Evo} approach we use the GVAE baseline with the latent space dimension $64$ together with the evolutionary operators presented in Section~\ref{sec:EA}. Here, different models find different equations: \textsc{GVAE 32} finds the equation \textit{NG-6}, while \textsc{GVAE 64} finds \textit{NG-4}, and \textsc{GVAE Evo} \textit{NG-3}. Overall \textsc{GVAE Evo} performs the best as it successfully finishes 4 runs more than other models. 

\begin{table}[h]
  \caption{Results of symbolic regression by random sampling and evolutionary optimization in the GVAE latent space with varying number of dimensions on the Nguyen benchark.}
  \label{tab:sr-gvae}
  \centering
  \resizebox{\linewidth}{!}{
    \begin{tabular}{lcccccccccccc}
    \toprule
          & \multicolumn{3}{l}{GVAE 32}                                             & \multicolumn{3}{l}{GVAE 64}                                   & \multicolumn{3}{l}{GVAE 128}                          & \multicolumn{3}{l}{GVAE Evo}\\
    Name  & Successful   & Mean $R^2$                   & Invalid                   & Successful   & Mean $R^2$               & Evaluated           & Successful & Mean $R^2$        & Invalid             & Successful & Mean $R^2$                   & Invalid        \\ \midrule
    NG-1  & 10           & 1.00 ($\pm$ 0.00)            & 15972 ($\pm$ 95)          & 9            & 1.00 ($\pm$ 0.01)        & 58229 ($\pm$ 149)   & 10         & 1.00 ($\pm$ 0.00) & 72863 ($\pm$ 96)    & 10         & 1.00 ($\pm$ 0.00)            & 0 ($\pm$ 1)    \\
    NG-2  & 2            & 1.00 ($\pm$ 0.01)            & 15917 ($\pm$ 111)         & 4            & 1.00 ($\pm$ 0.01)        & 58295 ($\pm$ 150)   & 3          & 1.00 ($\pm$ 0.01) & 72840 ($\pm$ 110)   & 7          & 1.00 ($\pm$ 0.01)            & 0 ($\pm$ 1)    \\
    NG-3  & 0            & 1.00 ($\pm$ 0.01)            & 15985 ($\pm$ 107)         & 0            & 1.00 ($\pm$ 0.01)        & 58383 ($\pm$ 153)   & 0          & 1.00 ($\pm$ 0.01) & 72802 ($\pm$ 101)   & 1          & 1.00 ($\pm$ 0.01)            & 1 ($\pm$ 0)    \\
    NG-4  & 0            & 1.00 ($\pm$ 0.01)            & 15900 ($\pm$ 115)         & 1            & 1.00 ($\pm$ 0.01)        & 58318 ($\pm$ 169)   & 0          & 1.00 ($\pm$ 0.01) & 72848 ($\pm$ 105)   & 0          & 1.00 ($\pm$ 0.01)            & 0 ($\pm$ 0)    \\
    NG-5  & 0            & 0.00 ($\pm$ 0.01)            & 15896 ($\pm$ 102)         & 0            & 0.00 ($\pm$ 0.01)        & 58322 ($\pm$ 137)   & 0          & 0.00 ($\pm$ 0.00) & 72808 ($\pm$ 97)    & 0          & 0.00 ($\pm$ 0.01)            & 0 ($\pm$ 1)    \\
    NG-6  & 2            & 0.59 ($\pm$ 0.20)            & 15925 ($\pm$ 120)         & 0            & 0.49 ($\pm$ 0.00)        & 58327 ($\pm$ 162)   & 0          & 0.49 ($\pm$ 0.00) & 72787 ($\pm$ 183)   & 0          & 0.53 ($\pm$ 0.09)            & 1 ($\pm$ 1)    \\
    NG-7  & 0            & 0.92 ($\pm$ 0.02)            & 15987 ($\pm$ 123)         & 0            & 0.92 ($\pm$ 0.00)        & 58268 ($\pm$ 107)   & 0          & 0.91 ($\pm$ 0.03) & 72865 ($\pm$ 116)   & 0          & 0.92 ($\pm$ 0.00)            & 0 ($\pm$ 1)    \\
    NG-8  & 10           & 1.00 ($\pm$ 0.00)            & 15986 ($\pm$ 132)         & 10           & 1.00 ($\pm$ 0.00)        & 58252 ($\pm$ 157)   & 10         & 1.00 ($\pm$ 0.00) & 72825 ($\pm$ 186)   & 10         & 1.00 ($\pm$ 0.00)            & 0 ($\pm$ 1)    \\\midrule
    Total/Mean  & 24     & 0.81 ($\pm$ 0.33)            & 15946 ($\pm$ 37)          & 24           & 0.81 ($\pm$ 0.34)        & 58299 ($\pm$ 45)    & 23         & 0.80 ($\pm$ 0.34) & 72830 ($\pm$ 27)    & 28         & 0.81 ($\pm$ 0.34)            & 0 ($\pm$ 0)    \\
    \bottomrule
\end{tabular}

  }
\end{table}

\section{Implementation details}\label{app:implementation}
This section provides implementation details that are not part of the methodology but are crucial for the reproducibility of our approach and its implementation. These details include batching, encoding/decoding, and training.

\subsection{Batching}
Since our approach works on expression trees with varying structures, we cannot employ standard batching methods. Instead, we represent a batch of expression trees with a Python object we refer to as a ``batched node’’. A batched node contains a list of symbols, a left (batched) child node, and a right (batched) child node. 

Since an expression tree may not contain all the nodes in the batched tree, the list of symbols within a batched node might include an empty string as a placeholder for the missing symbol. During training, each batched node also contains a target matrix, a prediction matrix, and a mask vector. The target matrix comprises one-hot symbol encodings, with empty strings represented by zero values. The decoding cell generates the prediction matrix, which, in turn, predicts the target matrix. Finally, the mask vector is a binary vector where the value at index $i$ equals one if the node appears in the expression tree $i$ and equals zero otherwise.

\subsection{Encoding/Decoding}
The encoding process involves traversing the batched node using a post-order traversal. We first visit and encode the left (batched) child, followed by the right (batched) child. Finally, we generate the code for the batched node using codes obtained from the child nodes.

Decoding is performed in reverse. We start by decoding the (batched) root code, which yields a list of symbols, the codes for the left and right (batched) child nodes, and the masks for the left and right child nodes. The masks for the left and right child nodes are calculated using symbols produced by the decoding cell and the mask of the current node. Specifically, if the value of the current mask at a given position is $0$, the corresponding values in the mask for the left and right child nodes remain $0$. Otherwise, the masks are assigned values appropriate to the symbol type: both $1$ for an operator, $1$, and right $0$ for a function, and both $0$ for a variable or constant. When all values in a mask become $0$, the decoding process for this branch terminates.

\subsection{Training}
Expression trees have two components: the binary tree structure and the symbols within the nodes. However, the structure of the binary tree can be inferred from the symbols present in the nodes. Therefore, it is sufficient for our approach to learn to reconstruct the symbols occurring in the nodes.

When training the model, we restrict the output tree's structure to match the input structure. Specifically, we utilize the input batched node and incorporate a prediction matrix into each node. We then calculate the reconstruction error using the target and prediction sequences obtained through an in-order traversal of the batched node. While computing the cross-entropy loss, we apply a masking technique to exclude nodes that do not occur in an expression tree from the loss calculation, effectively removing them from the training process.

\end{appendices}

\end{document}